\documentclass[10pt,twocolumn,letterpaper]{article}

\usepackage{cvpr}
\usepackage{times}
\usepackage[pdftex]{graphicx}
\usepackage{amsmath}
\usepackage{amssymb}
\usepackage{bm}
\usepackage{tabularx}
\usepackage{multirow}
\usepackage{algorithm}
\usepackage{algorithmic}

\usepackage{fmt}
\usepackage{mcr}
\usepackage{pkg}
\usepackage{thmE}
\usepackage{algE}

\usepackage[pagebackref=true,breaklinks=true,colorlinks,bookmarks=false]{hyperref}

\cvprfinalcopy 

\ifcvprfinal\fi
\begin{document}

\title{Multi-scale Domain-adversarial Multiple-instance CNN for Cancer Subtype Classification with Unannotated Histopathological Images
}

\author{
Noriaki Hashimoto$^1$\textsuperscript{$\dagger$},
Daisuke Fukushima$^1$\textsuperscript{$\dagger$},
Ryoichi Koga$^1$,
Yusuke Takagi$^1$,
Kaho Ko$^1$\\
Kei Kohno$^2$,
Masato Nakaguro$^2$,
Shigeo Nakamura$^2$,
Hidekata Hontani$^1$,
Ichiro Takeuchi$^{1,3}$\textsuperscript{*}\\
$^1$Nagoya Institute of Technology,
$^2$Nagoya University Hospital,
$^3$RIKEN\\
}

\maketitle
\thispagestyle{empty}

\begin{abstract}
We propose a new method for cancer subtype classification from histopathological images, which can automatically detect tumor-specific features in a given whole slide image (WSI).
The cancer subtype should be classified by referring to a WSI, i.e., a large-sized image (typically 40,000 $\times$ 40,000 pixels) of an entire pathological tissue slide, which consists of cancer and non-cancer portions.
One difficulty arises from the high cost associated with annotating tumor regions in WSIs.
Furthermore, both global and local image features must be extracted from the WSI by changing the magnifications of the image.
In addition, the image features should be stably detected against the differences of staining conditions among the hospitals/specimens.
In this paper, we develop a new CNN-based cancer subtype classification method by effectively combining multiple-instance, domain adversarial, and multi-scale learning frameworks in order to overcome these practical difficulties.
When the proposed method was applied to malignant lymphoma subtype classifications of 196 cases collected from multiple hospitals, the classification performance was significantly better than the standard CNN or other conventional methods, and the accuracy compared favorably with that of standard pathologists.

\end{abstract}
\footnote[0]{\textsuperscript{$\dagger$}N.H. and D.F. contributed equally.
\textsuperscript{*}Correspondence to I.T. (e-mail: takeuchi.ichiro@nitech.ac.jp).
}
\section{Introduction}
In this study, we propose a novel convolutional neural network (CNN)-based method for cancer subtype classification by using digital pathological images of hematoxylin-and-eosin (H\&E) stained tissue specimens as inputs.
Since a whole slide image (WSI) obtained by digitizing an entire pathological tissue slide is too large to feed into a CNN~{\cite{krizhevsky2012imagenet,simonyan2014very,he2016deep,chollet2017xception}, it is common to extract a large number of patches from the WSI~\cite{cirecsan2013mitosis,cruz2014automatic,mousavi2015automated,xu2015deep,hou2016patch,wang2016deep,bejnordi2017diagnostic,xu2017large,bandi2018detection,graham2018classification}.
If it can be known whether each patch is a tumor region or not, the CNN can be trained by using each patch as a labeled training instance.
However, the cost to annotate each of a large number of patch labels is too high.
When patch label annotation is not available, cancer subtype classification tasks are challenging in the following three respects.

The first difficulty is that tumor and non-tumor regions are mixed in a WSI.
Therefore, when pathologists actually conduct subtype classification, it is necessary to find out which part of the slide contains the tumor region, and perform subtype classification based on the features of the tumor region.
The second practical difficulty is that staining conditions vary greatly depending on the specimen conditions and the hospital from which the specimen was taken.
Therefore, pathologists perform tumor region identification and subtype classification by carefully considering the different staining conditions.
The last difficulty is that different features of tissues are observed when the magnification of the pathological image is changed.
Pathologists conduct diagnosis by changing the magnification of a microscope repeatedly to find out various features of the tissues.

In order to develop a practical CNN-based subtype classification system, our main idea is to introduce mechanisms that mimic these pathologist's actual practices.
To address the above three difficulties simultaneously, we effectively combine \emph{multiple instance learning (MIL)}, \emph{domain adversarial (DA) normalization}, and \emph{multi-scale (MS) learning} techniques.
Although each of these techniques has been studied in the literatures, we demonstrate that their effective and careful combination enables us to develop a CNN-based system
that performs significantly better than the standard CNN or other conventional methods.

We applied the proposed method to malignant lymphoma subtype classifications of 196 cases collected from 80 hospitals, and demonstrated that the accuracy of the proposed method compared favorably with standard pathologists.
It was also confirmed that the proposed method not only performed better than conventional methods, but also performed subtype classification in a similar way to pathologists in the sense that the method correctly paid attention to tumor regions in images of various different scales.

The main contributions of our study are as follows.
First, we developed a novel CNN-based digital pathology image classification method by effectively combining MIL, DA and MS approaches.
Second, we applied the proposed method to malignant lymphoma classification tasks with 196 WSIs of H\&E stained histological tissue slides, collected for the purpose of consultation by an expert pathologist on malignant lymphoma.
Finally, as a result of confirmation by immunostaining in the above malignant lymphoma subtype classification tasks, it was confirmed that the proposed method performed subtype classification by correctly paying attention to the true tumor regions from images at various different scales of magnification.

\section{Preliminaries}
Here we present our problem setup and three related techniques that are incorporated into the proposed method in the next section.
In this paper, we use the following notations.
For any natural number $N$, we define
$[N] := \{1, \ldots, N\}$.
We call a vector for which the elements are non-negative and sum-to-one a \emph{probability vector}.
Given two probability vectors $p, q$, $\cL(p, q)$ represents their cross entropy.

\subsection{Problem setup}
Consider a training set for a binary pathological image classification problem obtained from $N$ patients.
We denote the training set as
$
\{(\mathbb{X}_n, \mathbb{Y}_n)\}_{n=1}^N
$,
where
$\mathbb{X}_n$
is the whole slide image (WSI) and
$\mathbb{Y}_n$
is the two-dimensional class label one-hot vector
of the $n^{\rm th}$ patient
for
$n \in [N]$.
We also define
a set of $N$-dimensional vectors
$\{\mathbb{D}_n\}_{n=1}^N$
for which the $n^{\rm th}$ element is one and the others are zero.
Since each WSI is too huge to directly feed into a CNN, a patch-based approach is usually employed.
In this paper, we consider patches with $224 \times 224$ pixels.

In cancer pathology, since tumor and non-tumor regions are mixed, not all patches from a positive-class slide contain positive class-specific (tumor) information.
Thus, we borrow an idea from \emph{multiple instance learning (MIL)} (the detail of MIL will be described in \S~\ref{subsec:MIL}).
Specifically,
we consider a group of patches,
and assume that
each group from a positive class slide
contains
at least a few patches having positive class-specific information,
whereas
each group from a negative class slide
does not contain any
patches having positive-class specific information.
Furthermore,
when pathologists diagnose patients, they observe the glass slide at multiple different scales.
To mimic this, we consider patches with multiple different scales.

We denote the groups of patches at different scales as follows.
We use the notation $s \in [S]$ to indicate the index of scales (e.g., if scales 10x and 20x are considered, $S=2$).
The set of groups (called \emph{bags} in MIL framework) in the $n^{\rm th}$ WSI is denoted by $\cB_n$ for $n \in [N]$.
Then, each group (bag) $b \in \cB_n$ is characterized by a set of patches (called \emph{instances} in the MIL framework) $\cI_b^{(s)}$ for $b \in \cB_n$ and $s \in [S]$,
where
the superscript $^{(s)}$ indicates that
these patches are taken from scale $s$.
Figure~\ref{fig:GA} illustrates the notions of a WSI, groups (bags), patches (instances), and scales.

\begin{figure*}[tb]
\begin{center}
   \includegraphics[width=0.75\linewidth]{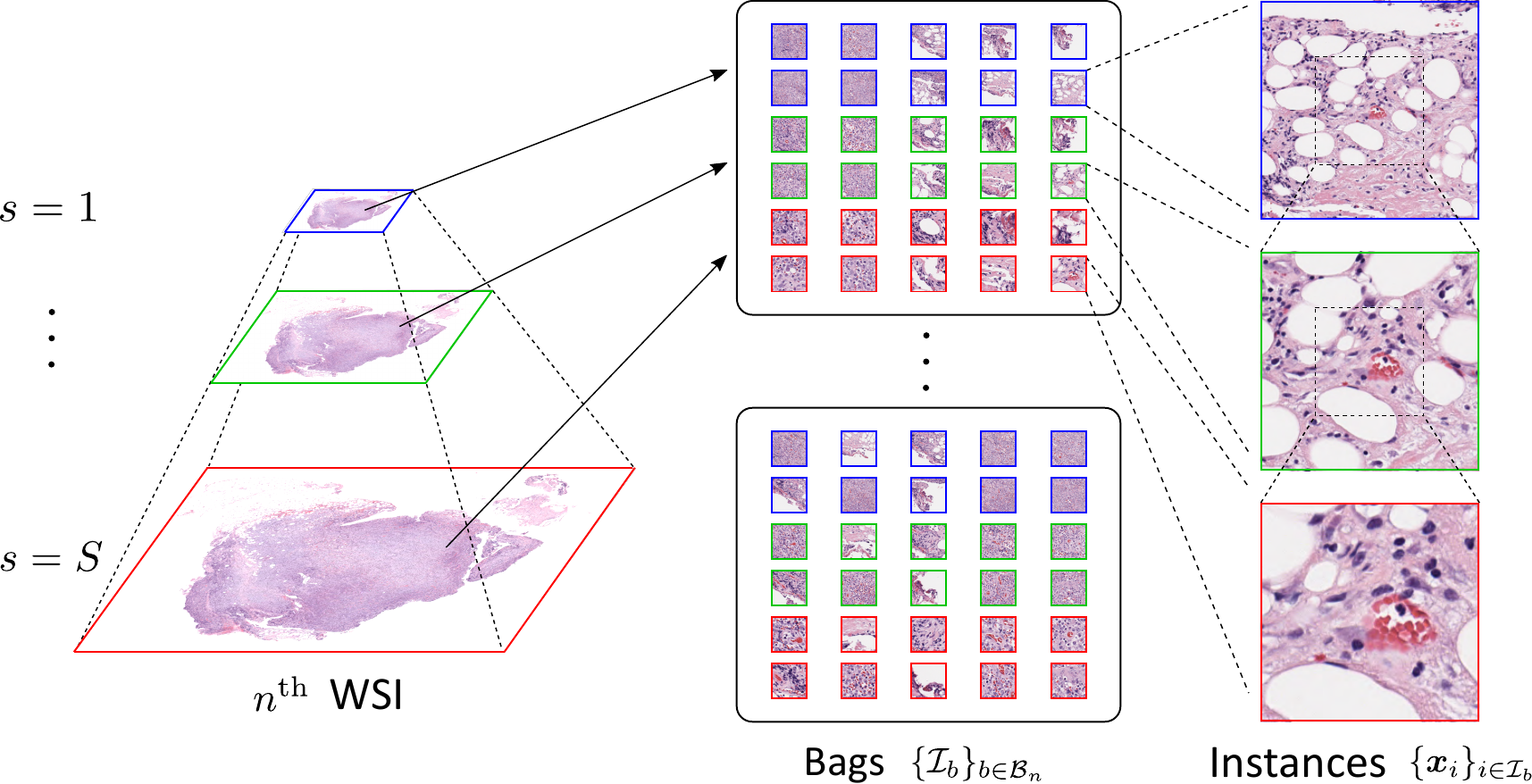}
\end{center}
 \caption{
 A brief illustration of the notions of a WSI, bags, instances (patches), and scales.
 A large number of $224 \times 224$-pixel image patches are extracted from an entire WSI at multiple different scales.
 In the problem setup considered in this paper, instance class labels are not observed, but the class labels for groups of instances called bags are observed.
 It is important to note that each bag contains patches taken at multiple different scales.
 This enables us to detect multiple regions of interest from multiple different scale images.
}
 \label{fig:GA}
\end{figure*}

\subsection{Multiple instance learning (MIL)}
\label{subsec:MIL}
Multiple-instance learning (MIL) is a type of weakly supervised learning problem, where instance labels are not observed but labels for groups of instances called \emph{bags} are observed.
In the binary classification setting, a positive label is assigned to a bag if the bag contains at least one positive instance, while a negative label is assigned to a bag if the bag only contains negative instances.
Figure~\ref{fig:MIL} illustrates MIL for a binary classification problem.
Various models and learning algorithms for MIL have been studied in the literatures~{\cite{dietterich1997solving,maron1998framework,zhou2002neural,andrews2003support,li2012efficient,chen2013multi,wu2015deep}.

MIL has been successfully applied to classification problems with histopathological images~\cite{cosatto2013automated,das2018multiple,ilse2018attention,couture2018multiple,sudharshan2019multiple,campanella2019clinical}.
For example, for binary classification of malignant and benign patients, WSIs for malignant patients contain both malignant and benign patches, while WSIs for benign patients only contain benign patches.
If we regard the WSIs for malignant/benign patients as positive/negative bags and malignant/benign patches as positive/negative instances, respectively, the above binary classification problem can be interpreted as an MIL problem.
The MIL framework is useful in histopathological image classification when no annotation is made for each extracted patch.
Our main idea in this paper is to use MIL framework in order to automatically identify multiple regions of interest at multiple different scales.
\begin{figure}[t]
\begin{center}
   \includegraphics[width=0.8\linewidth]{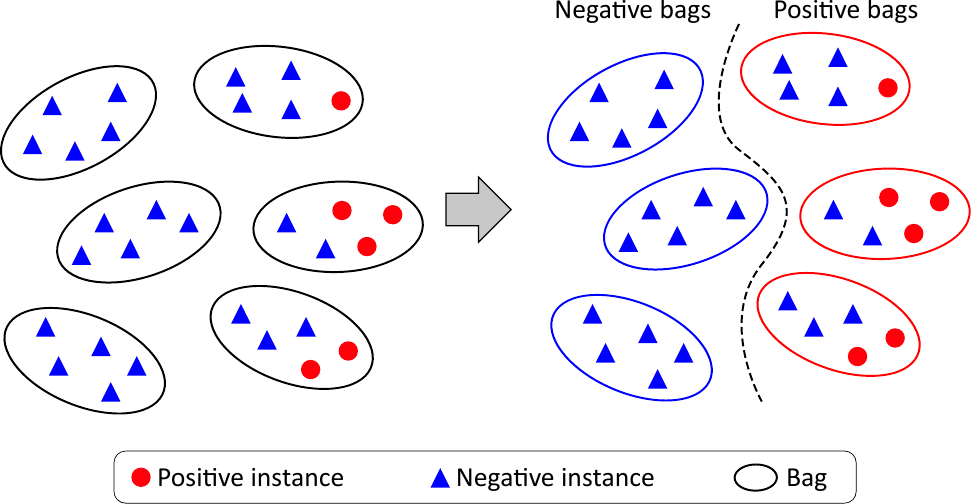}
\end{center}
   \caption{Explanation of MIL. Positive bags are generated from WSIs with positive subtype labels and negative bags are generated from WSIs with negative subtype labels. Only the image patches of class-specific regions, such as tumors in positive-class WSIs, are regarded as positive instances.}
\label{fig:MIL}
\end{figure}
\vspace{-3mm}
\subsection{Domain-adversarial neural network}
Slide-wise differences in staining conditions,
as illustrated in Fig.~\ref{fig:stain},
highly degrade
the classification accuracy.
To overcome this difficulty,
appropriate pre-processing
such as color normalization~\cite{reinhard2001color,macenko2009method,khan2014nonlinear,bejnordi2015stain,zanjani2018histopathology} or
color augmentation~\cite{lafarge2017domain,rakhlin2018deep}
would be required.
Color normalization adjusts the color of input images to the target color distribution.
Color augmentation suppresses the effect of outlying colors by generating augmented images while slightly changing the color of an original image.

Domain-adversarial (DA)~\cite{ganin2016domain} training has been proposed to ignore the differences among training instances that do not contribute to the classification task.
In the histopathological image classification setting,
Lafarge et al.~\cite{lafarge2017domain}
introduced a DA training approach,
and demonstrated that it was superior to color augmentation, stain normalization, and their combination.
In the proposed method,
we use a DA training approach within the MIL framework
for histopathological image classification
by regarding
each patient as an individual domain
so that
the staining condition of each patient's slide
can effectively be ignored.

\begin{figure}[t]
  \begin{center}
  \begin{tabular}{ccc}
    \begin{minipage}[t]{0.27\hsize}
      \centering
      \includegraphics[width=\hsize]{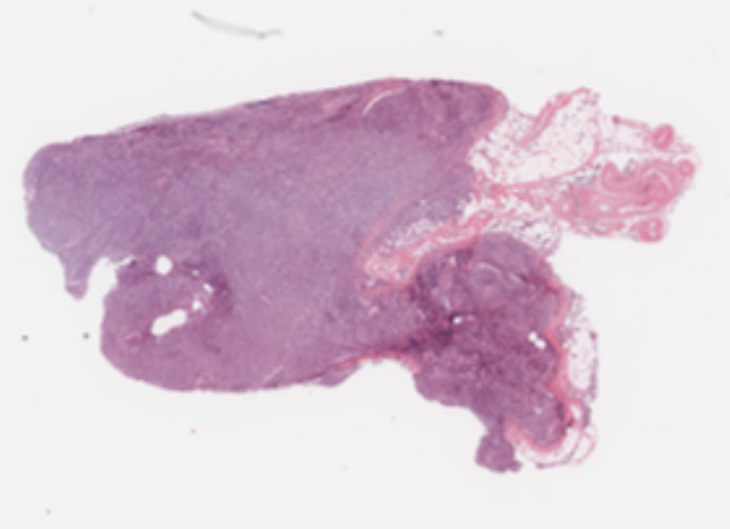}
    \end{minipage} &
    \begin{minipage}[t]{0.27\hsize}
      \centering
      \includegraphics[width=\hsize]{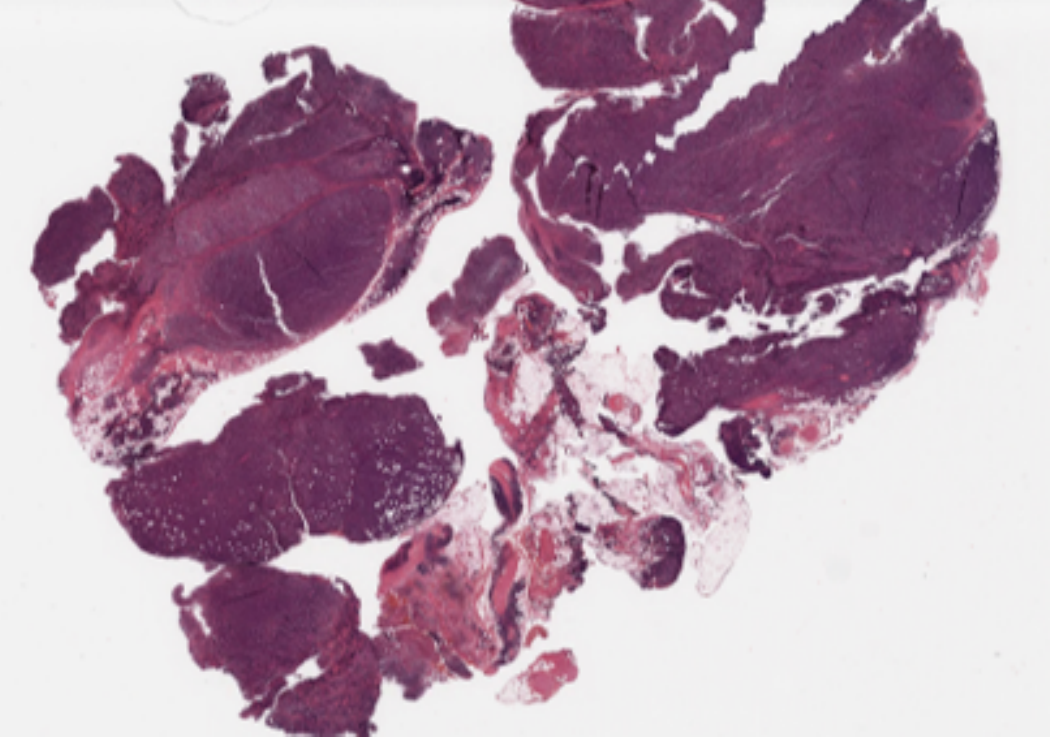}
    \end{minipage} &
    \begin{minipage}[t]{0.27\hsize}
      \centering
      \includegraphics[width=\hsize]{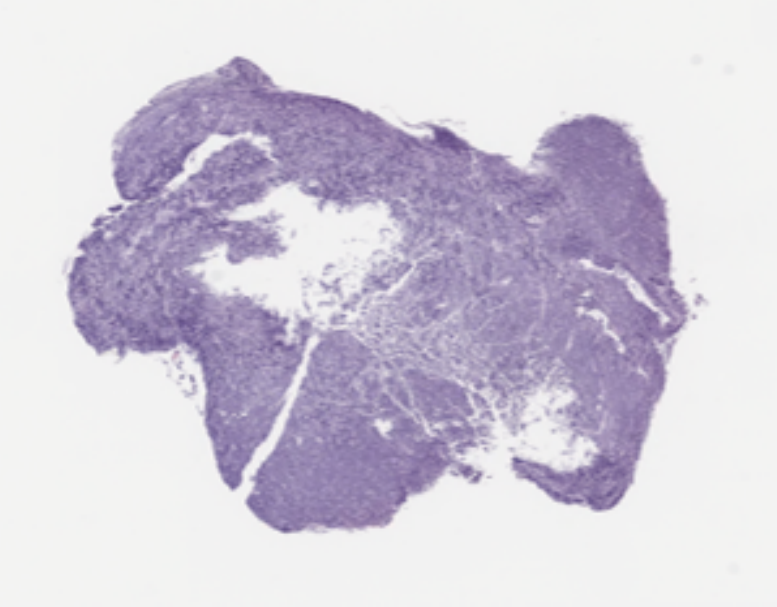}
    \end{minipage}
  \end{tabular}
  \begin{tabular}{ccc}
    \begin{minipage}[t]{0.27\hsize}
      \centering
      \includegraphics[width=\hsize]{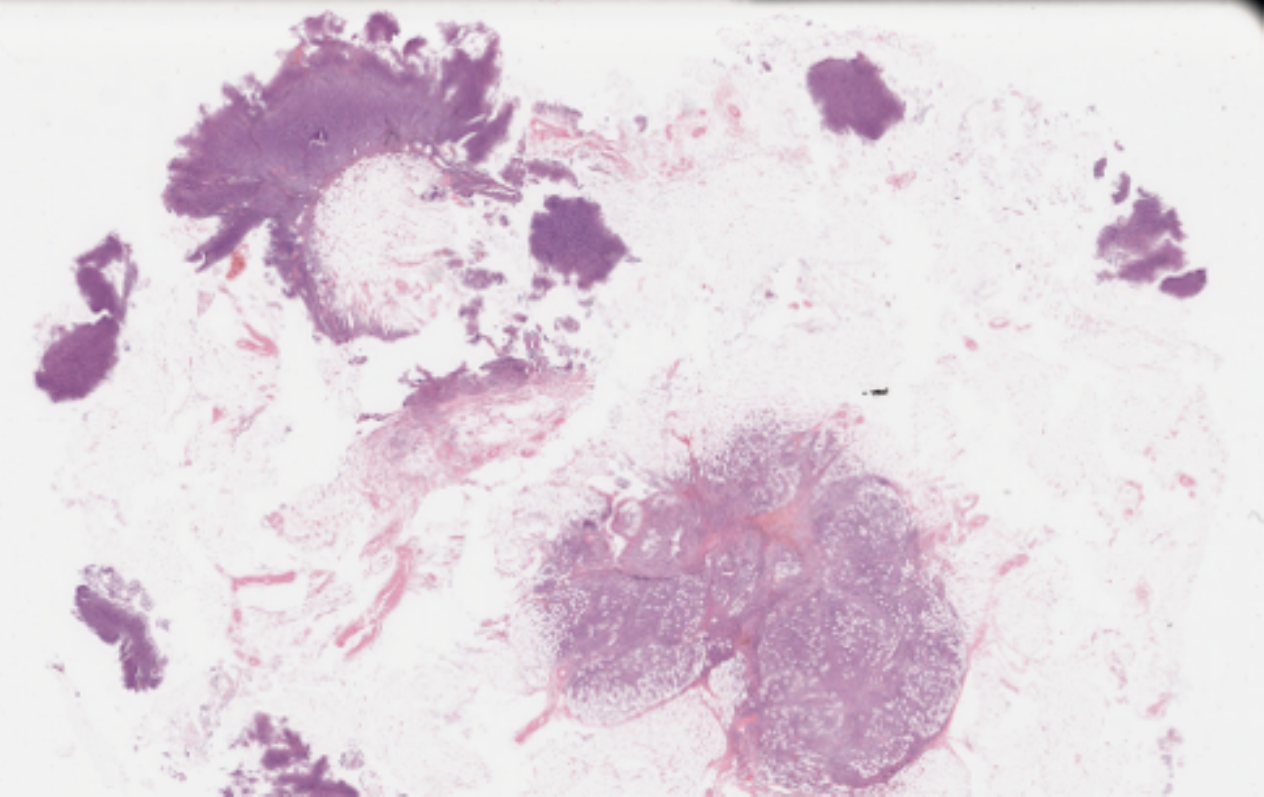}
    \end{minipage} &
    \begin{minipage}[t]{0.27\hsize}
      \centering
      \includegraphics[width=\hsize]{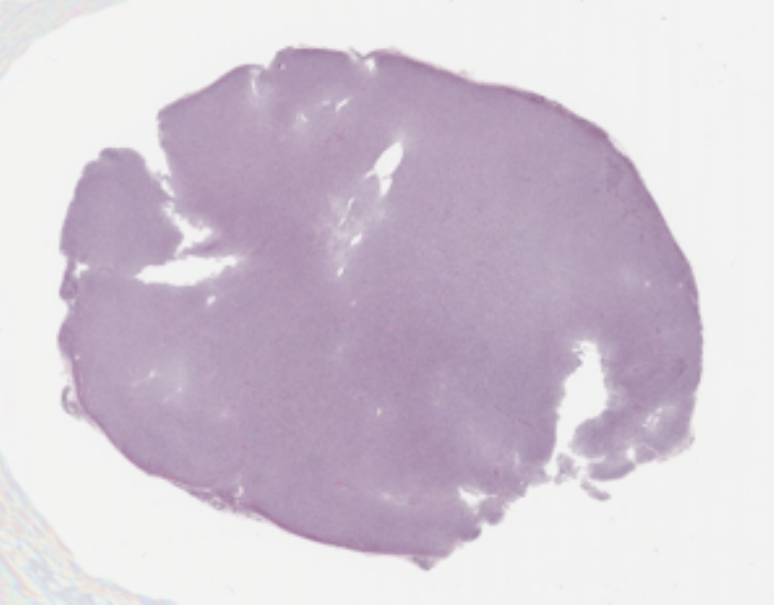}
    \end{minipage} &
    \begin{minipage}[t]{0.27\hsize}
      \centering
      \includegraphics[width=\hsize]{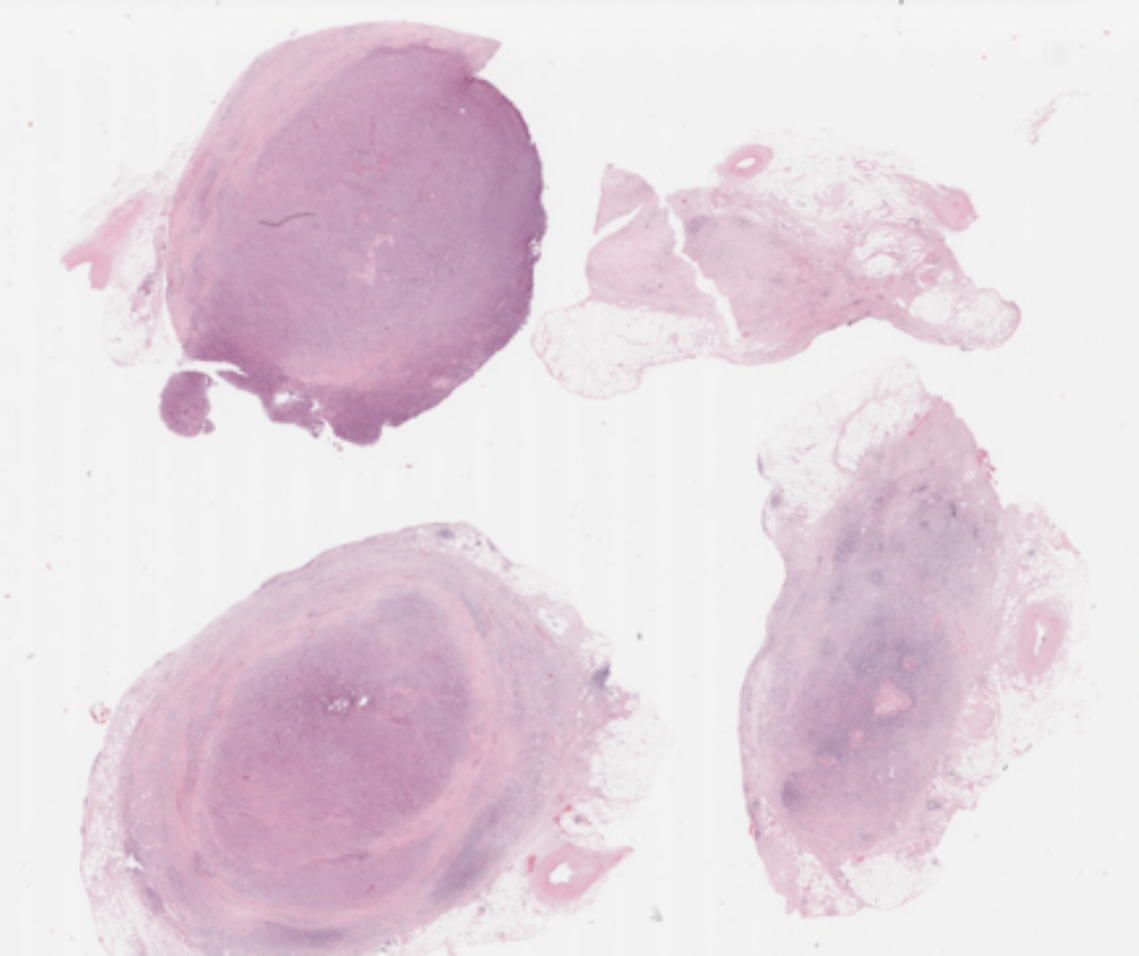}
    \end{minipage}
  \end{tabular}
  \begin{tabular}{ccc}
    \begin{minipage}[t]{0.27\hsize}
      \centering
      \includegraphics[width=\hsize]{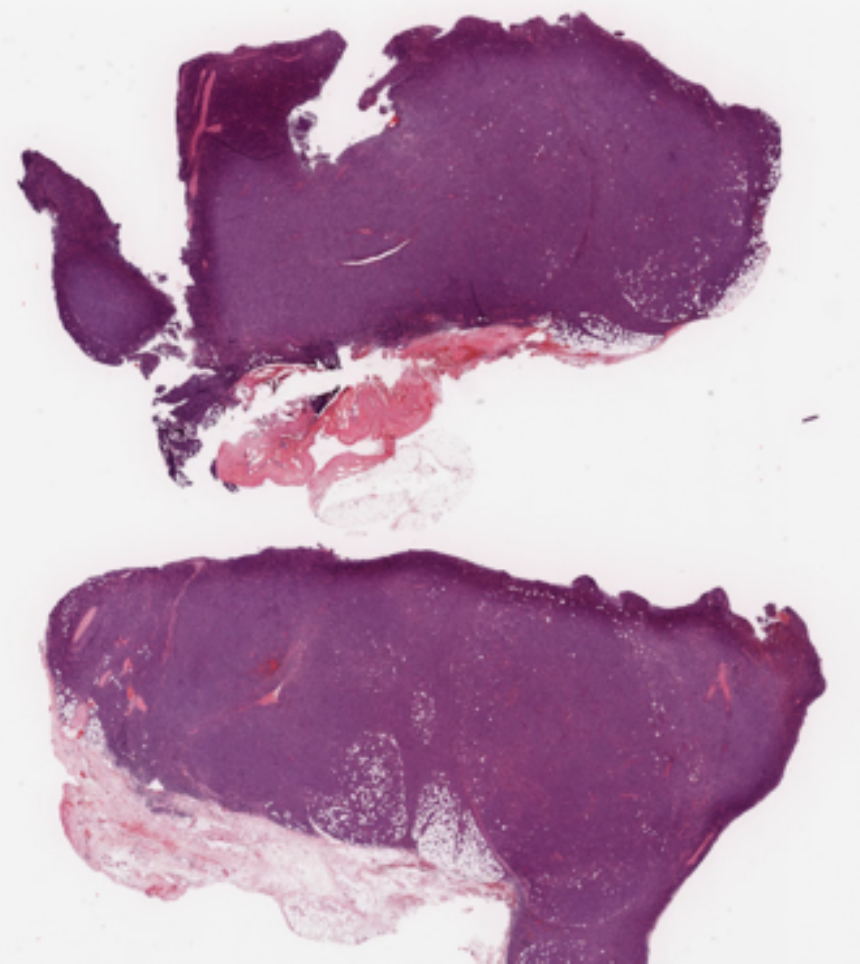}
    \end{minipage} &
    \begin{minipage}[t]{0.27\hsize}
      \centering
      \includegraphics[width=\hsize]{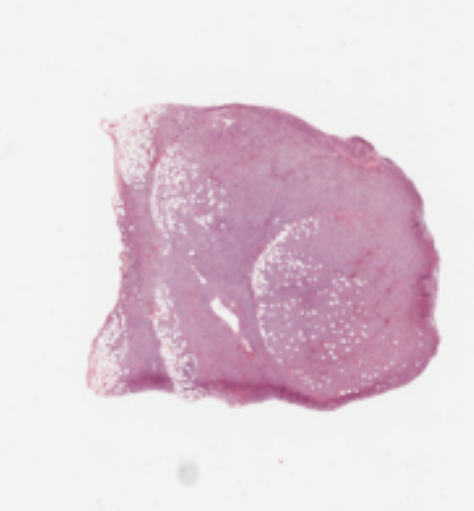}
    \end{minipage} &
    \begin{minipage}[t]{0.27\hsize}
      \centering
      \includegraphics[width=\hsize]{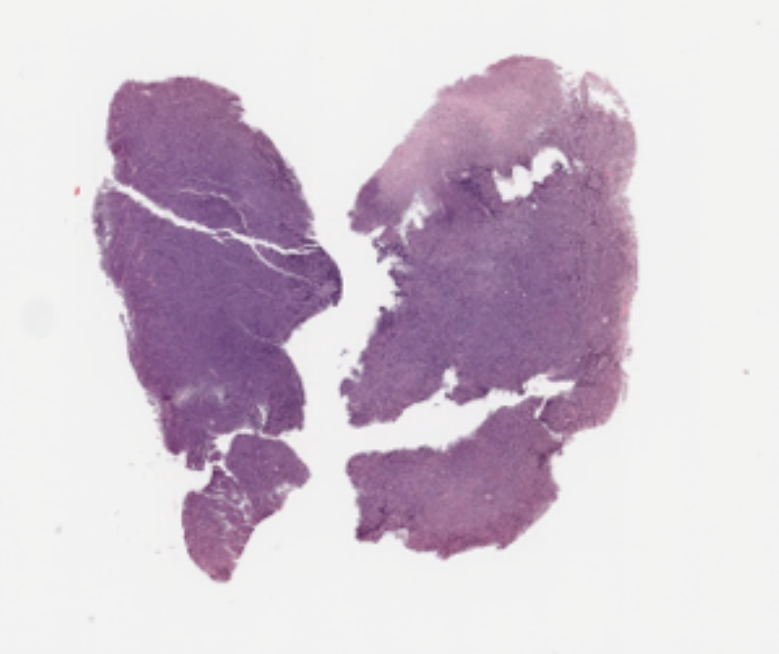}
    \end{minipage}
  \end{tabular}
\end{center}
  \caption{Entire WSIs of H\&E stained tissues prepared at different facilities. Variety in staining conditions can be seen among different staining protocols.}
  \label{fig:stain}
\end{figure}

\subsection{Multi-scale pathology image analysis}
Pathologists observe different features at different scales of magnification.
For example, global tissue structure and detailed shapes of nuclei can be seen at low and high scales of magnification, respectively.
Although most of the existing studies on histopathological image analysis use a fixed single scale, some studies use multiple scales~\cite{bejnordi2015multi,gao2016multi,wetteland2019multiscale,tokunaga2019adaptive}.

When multi-scale images are available in histopathological image analysis, a common approach is to use them \emph{hierarchically} from low resolution to high resolution.
Namely, a low-resolution image is first used to detect regions of interest, and then high-resolution images of the detected regions are used for further detailed analysis.
Another approach is to automatically \emph{select} the appropriate scale from the image itself.
For example, Tokunaga et al.~\cite{tokunaga2019adaptive} employed a mixture-of-expert network, where each expert was trained with images of different scale, and the gating network selected which expert should be used for segmentation.

In this study, we noted that expert pathologists conduct diagnosis by changing the magnification of a microscope repeatedly to find out various features of the tissues.
This indicates that the analysis of multiple regions at multiple scales plays a significant role in subtype classification.
In order to mimic this pathologists' practice, we propose a novel method to use multiple patches at multiple different scales within the MIL framework.
In contrast to the hierarchical or selective usage of multi-scale images,
our approach uses multi-scale images simultaneously.

\begin{figure*}[t]
\begin{center}
   \includegraphics[width=0.75\linewidth]{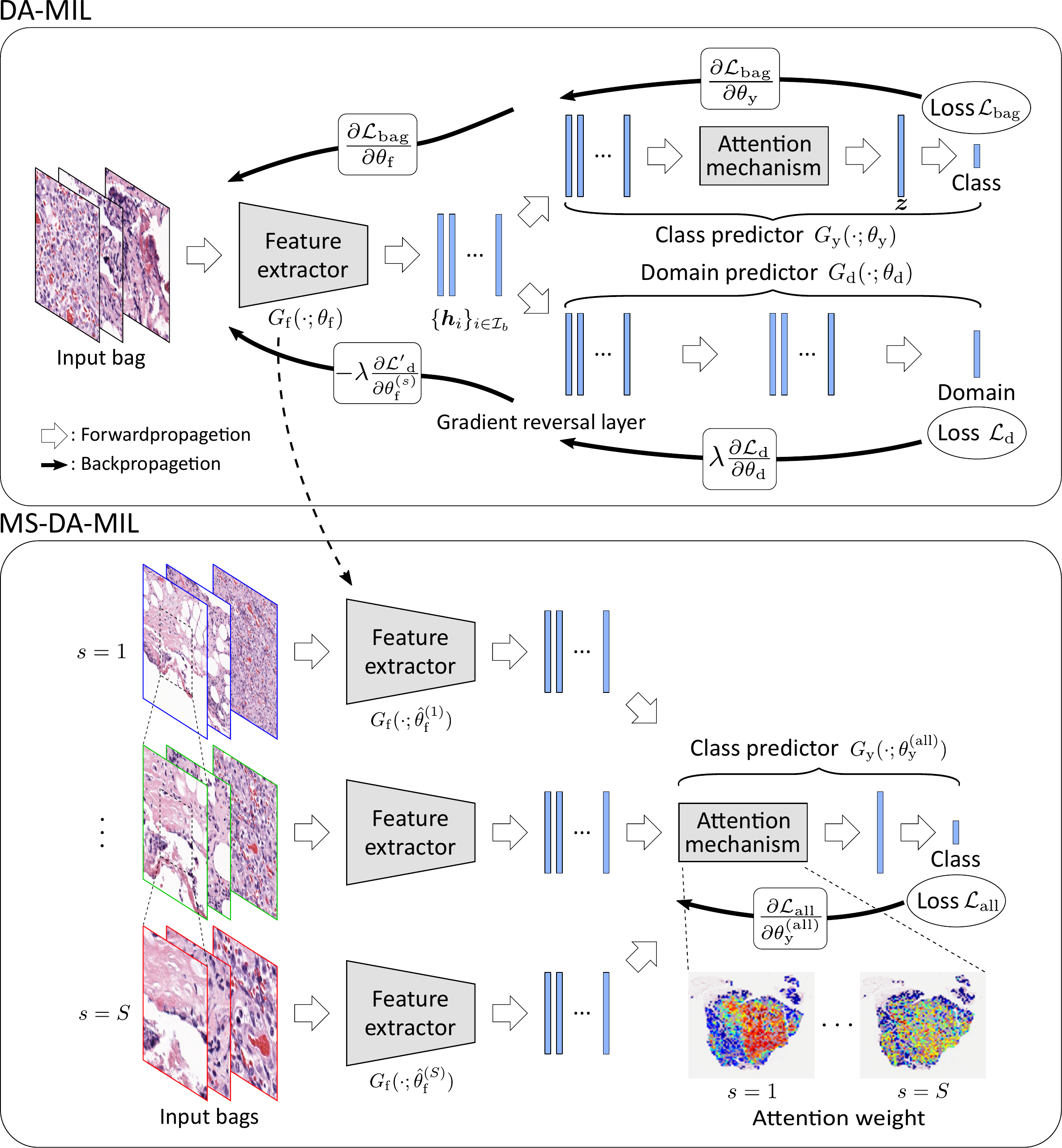}
\end{center}
 \caption{
 An illustration of the structure of the proposed network.
 Single scale DA-MIL networks are trained in stage 1 for each scale $s \in [S]$ (top).
 A multi-scale DA-MIL (MS-DA-MIL) network is trained in stage 2 (bottom).
 Loss function $\mathcal{L}_{\rm{bag}}$ and $\mathcal{L}_{\rm d}$ are the loss functions for the predicted bag labels and domain labels in eq.~\eq{eq:problem_s}.
 In MS-DA-MIL, feature extractors $G_{\rm{f}}^{(s)}$, which were domain-adversarially trained with DA-MIL are employed to generate feature vectors from the instances in bags $\cI_{b}^{(s)}$ and those feature vectors for all $S$ scales are aggregated for calculating attention weights.
 }
\label{fig:model}
\end{figure*}

\section{Proposed method}
In the proposed method,
the subtype of each patient
is predicted based on the H\&E stained WSI
by summarizing the predicted class labels
of the bags taken from the WSI.
Specifically,
given a test WSI
$\mathbb{X}_n$,
the class label probability
is simply predicted as
$P(\hat{\mathbb{Y}}_n = 1) = p_1/(p_1 + p_0)$,
where
\begin{align*}
 p_1 = \exp
 \left(\frac{1}{|\cB_n|}
 \sum_{b \in \cB_n}
 \log P(\hat{Y}_{b} = 1)
 \right),\\
 p_0 = \exp
  \left(\frac{1}{|\cB_n|}
  \sum_{b \in \cB_n}
  \log P(\hat{Y}_{b} = 0)
  \right).
\end{align*}
Here,
$P(\hat{Y}_b = 1)$
and
$P(\hat{Y}_b = 0)$
are the class label probabilities of the bag
$b \in \cB_n$.

The bag's class label probability
is
obtained as the output of the proposed CNN network, as depicted in Fig.~\red{\ref{fig:model}}.
It consists of the following three building blocks.
Feature extractor
      $G_{\rm{f}}: \bm{x} \mapsto \bm{h}$
      is a CNN which maps a $224 \times 224$-pixel image $\bm{x}$ into a $Q$-dimensional feature vector $\bm{h}$.
      It is denoted as
      $\bm{h} = G_{\rm{f}}(\bm{x} ; \theta_{\rm{f}})$
      where
      $\theta_{\rm{f}}$
      is the set of trainable parameters.
      Bag class label predictor
      $G_{\rm{y}}: \{\bm{h}_i\}_{i \in \cI_b} \mapsto P(\hat{Y}_{b})$
      is an NN with an attention mechanism~\cite{ilse2018attention}
      that maps the set of feature vectors in a bag $b$ into the probabilities of the bag class label $\hat{Y}_b$.
      $G_{\rm{y}}(\cdot ; \theta_{\rm{y}})$ is characterized by a set of trainable parameters $\theta_{\rm{y}}$, where $(\bm{V}, \bm{w}) \in \theta_{\rm{y}}$ are the sets of parameters for the attention network.
      Using $Q^{\prime}$-dimensional feature vectors $\{\bm{h}^{\prime}_i\}_{i \in \cI_b}$ generated through the fully connected layer,
      the attention weighted feature vector $\bm{z} \in \RR^{Q^{\prime}}$ is obtained as
      $\bm{z} = \sum_{i \in \cI_b} a_i \bm{h}^{\prime}_i$,
      where each attention is defined as
      \begin{align*}
       a_i = \frac{
       \exp
       \left(
       \bm{w}^\top {\rm tanh}(\bm{V} \bm{h}^{\prime}_i)
       \right)
       }{
       \sum_{j \in \cI_b}
       \exp
       \left(
       \bm{w}^\top {\rm tanh}(\bm{V} \bm{h}^{\prime}_{j})
       \right)
       },
       i \in \cI_b.
      \end{align*}
      Domain predictor
      $G_{\rm{d}}: \bm{h} \mapsto P(\hat{d})$
      is a simple NN
      that maps a feature vector $\bm{h}$ into domain label probabilities $P(\hat{d})$.
      It is denoted as
      $G_{\rm{d}}(\bm{h};\theta_{\rm{d}})$,
      where
      $\theta_{\rm{d}}$
      is the set of trainable parameters.
Training of the proposed CNN network is conducted in two stages.
In the first stage,
a single-scale DA-MIL network (the top one in Fig.~\red{\ref{fig:model}})
is trained
to obtain
the feature extractor
$G_{\rm{f}}(\cdot ; \theta_{\rm{f}}^{(s)})$
for each scale
$s \in [S]$.
Then,
in the second stage,
a multi-scale DA-MIL network (the bottom one in Fig.~\red{\ref{fig:model}})
is trained by plugging the $S$ trained feature extractors into the network.

\subsection{Stage1: single-scale learning}
In stage 1,
a single-scale DA-MIL network
is trained
for each scale $s \in [S]$
to predict the bag class labels
where
each bag only contains patches from the image of scale $s$.
Here
we modified the DA regularization~\cite{ganin2016domain} in order to apply it to only image patches with lower attention weights of MIL.
The training task of a single-scale DA-MIL network is formulated as the following minimization problem:
\begin{align}
 \nonumber
 &
 \left(
 \hat{\theta}_{\rm{f}}^{(s)},
 \hat{\theta}_{\rm{y}}^{(s)},
 \hat{\theta}_{\rm{d}}^{(s)}
 \right)
 \leftarrow
 \arg \min_{
 \theta_{\rm{f}}^{(s)},
 \theta_{\rm{y}}^{(s)},
 \theta_{\rm{d}}^{(s)}
 }
 \sum_{n = 1}^N
 \sum_{b \in \cB_n}
 \cL(\mathbb{Y}_n, P(\hat{Y}_{b}^{(s)}))
 \\
 \label{eq:problem_s}
 &
 -
 \lambda
 \sum_{n = 1}^N
 \sum_{b \in \cB_n}
 \frac{1}{|\cI_{b}^{(s)}|}
 \sum_{i \in \cI_{b}^{(s)}}
 \beta_i
 \cL(\mathbb{D}_n, G_{\rm{d}}(\bm{h}_i;\theta_{\rm{d}}^{(s)})),
\end{align}
where
\begin{align*}
   P(\hat{Y}_{b}^{(s)}) &= G_{\rm{y}} \left( \{G_{\rm{f}}(\bm{x}_{i}; \theta_{\rm{f}}^{(s)})\}_{i \in \cI_{b}^{(s)}}; \theta_{\rm{y}}^{(s)} \right),
   \\
   \beta_i &= \max_{a_j}\{a_j|j \in \cI_{b}^{(s)}\}-a_i.
\end{align*}
In eq.~\eq{eq:problem_s},
the first term is
the loss function for bag class label prediction,
while
the second term
is the penalty function for DA regularization, which is weighted by attentions for each instance.
The loss function is simply defined by the cross entropy
between
the true class label
and
predicted class label probability.
Here,
the bag class label is predicted
by only using instances for which the attentions are large.
The DA regularization term
is also defined by the cross entropy
between
the domain label
and
the predicted domain label probability.
By penalizing the domain prediction capability using DA regularization,
the feature extractor
$G_{\rm{f}}(\cdot ; \theta_{\rm{f}}^{(s)})$
for
each
$s \in [S]$
is trained so that the difference in staining conditions can be ignored.

\subsection{Stage2: multi-scale learning}
In stage 2,
a multi-scale DA-MIL network
is trained
to predict the bag class label
where
each bag contains instances (patches) across different scales.
The bag class label is predicted as
\begin{align*}
 P(\hat{Y}_{b})
 =
 G_{\rm{y}} \left( \{\{G_{\rm{f}}(\bm{x}_{i}; \hat{\theta}_{\rm{f}}^{(s)})\}_{i \in \cI_{b}^{(s)}}\}_{s=1}^S; \theta_{\rm{y}}^{(\rm all)} \right),
\end{align*}
where
the set of feature extractors
$G_{\rm{f}}(\cdot; \hat{\theta}_{\rm{f}}^{(s)})$,
$s \in [S]$,
which were already trained in the first stage, are plugged in.
The training of the set of parameters $\theta_{\rm{y}}^{(\rm all)}$ is formulated as the following minimization problem:
\begin{align}
 \label{eq:problem_all}
 \hat{\theta}_{\rm{y}}^{(\rm all)} \leftarrow \arg \min_{\theta_{\rm{y}}^{(\rm all)}}
 \sum_{n=1}^N
 \sum_{b \in \cB_n}
 \cL(\mathbb{Y}_n, P(\hat{Y}_{b})).
\end{align}

\subsection{Algorithm}

The algorithm of our proposed method is described in Algorithm \ref{algo:MS-MIL}.
Each parameter update is conducted by using the instances (patches) in each bag as a mini-batch.

\renewcommand{\algorithmicrequire}{\textbf{Input:}}
\renewcommand{\algorithmicensure}{\textbf{Output:}}
\begin{algorithm} [t]
    \caption{Parameter update in MS-DA-MIL training.}
    \label{algo:MS-MIL}
    \begin{algorithmic}
    \REQUIRE training set $\{(\mathbb{X}_n, \mathbb{Y}_n)\}_{n=1}^N$ with domain label $\{\mathbb{D}_n\}_{n=1}^N$, learning rate $\eta$, domain regularization parameter $\lambda$, train epochs $M$
    \\
    \% {\bf stage 1}: train feature extractor $G_{\rm{f}}(\cdot;\theta_{\rm{f}}^{(s)})$, class predictor $G_{\rm{y}}(\cdot;\theta_{\rm{y}}^{(s)})$, domain predictor $G_{\rm{d}}(\cdot;\theta_{\rm{d}}^{(s)})$
    \FOR {$m=1$ to $M$}
    \FOR {$s=1$ to $S$}
    \FOR {$n=1$ to $N$}
    \FOR {$b=1$ to $|\mathcal{B}_n|$}
        \STATE $\{\bm{h}_i\}_{i \in \mathcal{I}_{b}^{(s)}} \lA \{G_{\rm{f}}(\bm{x}_{i} ; \theta_{\rm{f}}^{(s)})\}_{i \in \mathcal{I}_{b}^{(s)}}$
        \STATE $\mathcal{L}_{\rm{bag}} \lA \mathcal{L}\left(\mathbb{Y}_n, G_{\rm{y}}(\{\bm{h}_i\}_{i \in \mathcal{I}_{b}^{(s)}}; \theta_{\rm{y}}^{(s)})\right)$
        \STATE $\mathcal{L}_{\rm{d}} \lA \frac{1}{|\mathcal{I}_{b}^{(s)}|} \sum_{i \in \mathcal{I}_{b}^{(s)}}
        \mathcal{L}\left(\mathbb{D}_{n},G_{\rm{d}}(\bm{h}_i ; \theta_{\rm{d}}^{(s)})\right)$
        \STATE $\mathcal{L^{\prime}}_{\rm{d}} \lA \frac{1}{|\mathcal{I}_{b}^{(s)}|} \sum_{i \in \mathcal{I}_{b}^{(s)}}
        \beta_i
        \mathcal{L}\left(\mathbb{D}_{n},G_{\rm{d}}(\bm{h}_i; \theta_{\rm{d}}^{(s)})\right)$
        \STATE $\beta_i = \max_{a_j}\{a_j|j \in \cI_{b}^{(s)}\}-a_i$
        \STATE $\theta_{\rm{y}}^{(s)} \leftarrow \theta_{\rm{y}}^{(s)}- \eta \frac{\partial \mathcal{L}_{\rm{bag}}}{\partial \theta_{\rm{y}}^{(s)}}$
        \STATE $\theta_{\rm{d}}^{(s)} \leftarrow \theta_{\rm{d}}^{(s)} - \eta \lambda \frac{\partial \mathcal{L}_{\rm{d}}}{\partial \theta_{\rm{d}}^{(s)}} $
        \STATE $\theta_{\rm{f}}^{(s)} \leftarrow \theta_{\rm{f}}^{(s)} - \eta\left(\frac{\partial \mathcal{L}_{\rm{bag}}}{\partial \theta_{\rm{f}}^{(s)}} - \lambda \frac{\partial \mathcal{L^{\prime}}_{\rm{d}}}{\partial \theta_{\rm{f}}^{(s)}} \right)$
    \ENDFOR\\
    \ENDFOR\\
    \ENDFOR\\
    \ENDFOR\\
    \% {\bf stage 2}: train class predictor $G_{\rm{y}}(\cdot; \theta_{\rm{y}}^{(\rm{all})})$
    \FOR {$m=1$ to $M$}
    \FOR {$n=1$ to $N$}
    \FOR {$b=1$ to $|\mathcal{B}_n|$}
    \STATE $\cL_{\rm all} \lA \cL(\mathbb{Y}_n, P(\hat{Y}_b))$
    \STATE $P(\hat{Y}_{b})=G_{\rm{y}}(\{\{G_{\rm{f}}(\bm{x}_{i} ; \theta_{\rm{f}}^{(s)})\}_{i \in \mathcal{I}_{b}^{(s)}}\}_{s=1}^{S} ; \theta_{\rm{y}}^{(\rm{all})})$
    \STATE $\theta_{\rm{y}}^{(\rm{all})} \leftarrow \theta_{\rm{y}}^{\rm{(all)}} - \eta \frac{\partial \mathcal{L}_{\rm all}}{\partial \theta_{\rm{y}}^{\rm{(all)}}}$
    \ENDFOR
    \ENDFOR
    \ENDFOR
    \ENSURE neural network $\{\{\theta_{\rm{f}}^{(s)}\}_{s=1}^S, \theta_{\rm{y}}^{\rm{(all)}}\}$
    \end{algorithmic}
\end{algorithm}

\begin{figure*}[h]
  \begin{center}
  \begin{tabular}{ccc}
    \begin{minipage}[t]{0.235\hsize}
      \centering
      \includegraphics[width=1.0\hsize]{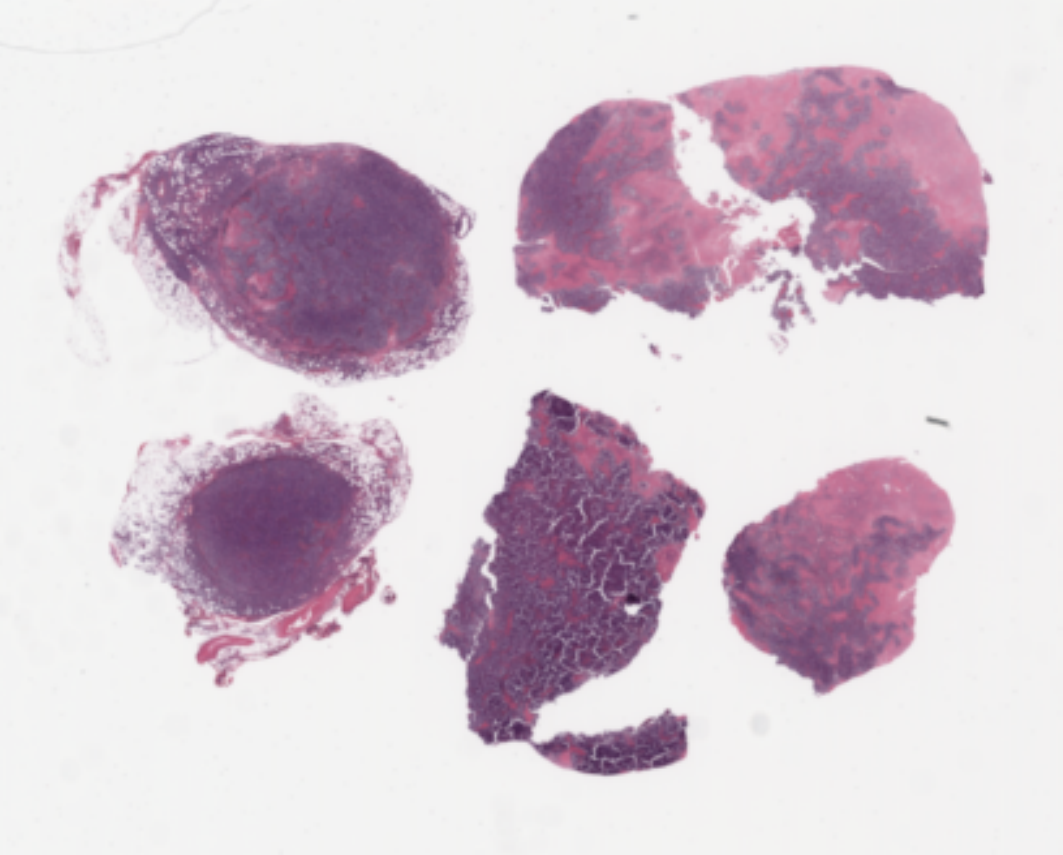}
    \end{minipage} &
    \begin{minipage}[t]{0.235\hsize}
      \centering
      \includegraphics[width=1.0\hsize]{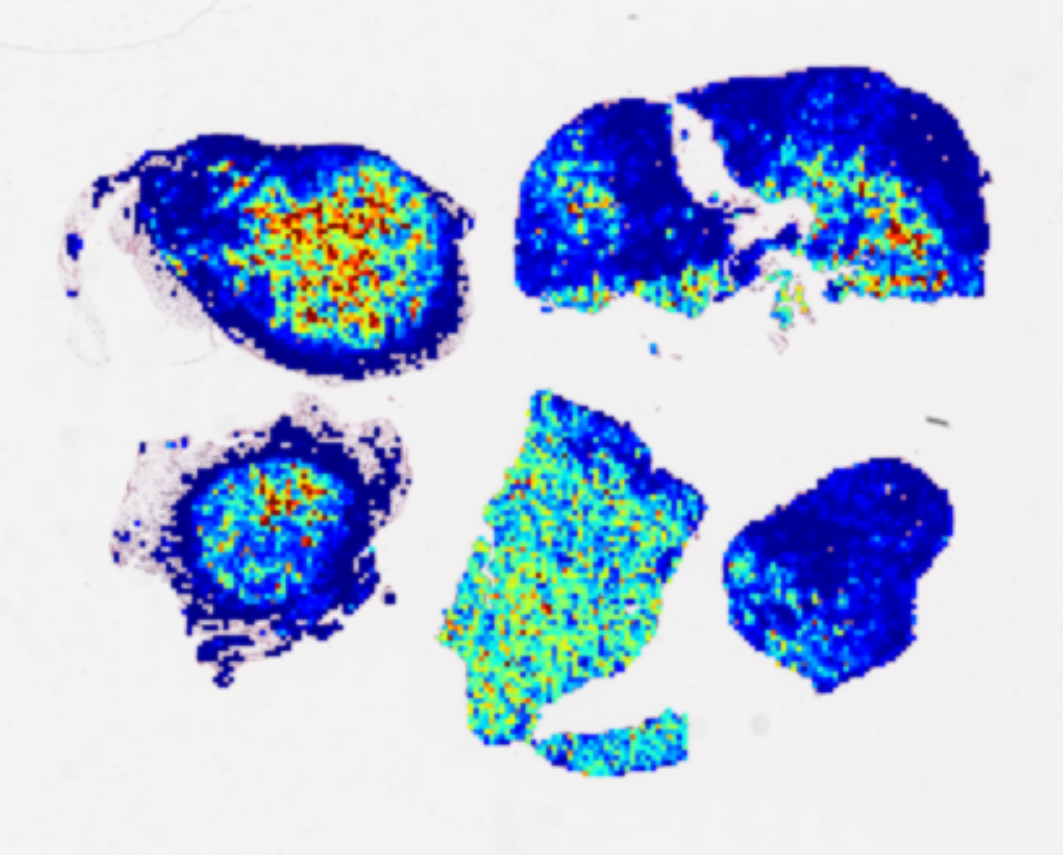}
    \end{minipage} &
    \begin{minipage}[t]{0.235\hsize}
      \centering
      \includegraphics[width=1.0\hsize]{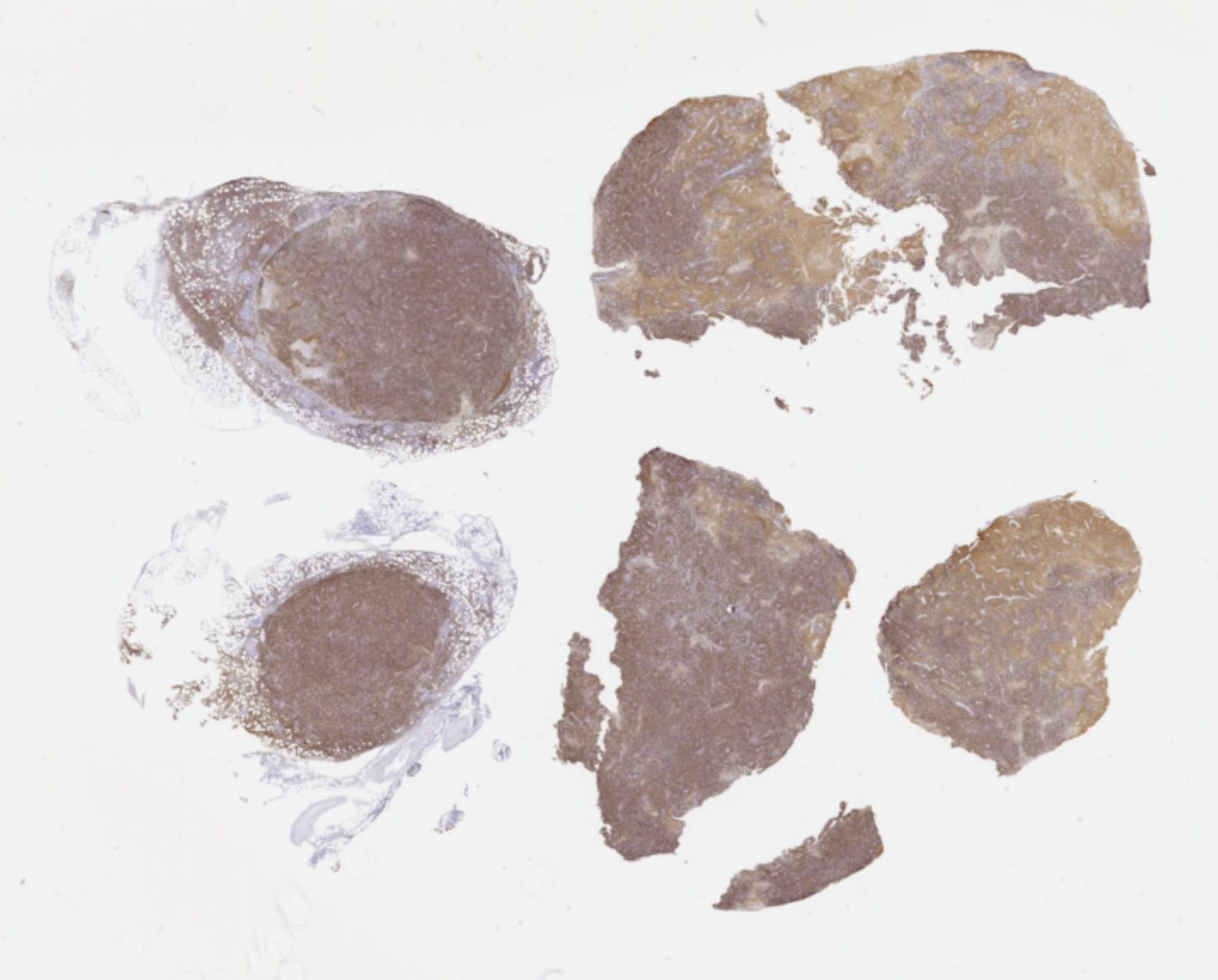}
    \end{minipage}
  \end{tabular}
  \begin{tabular}{ccc}
    \begin{minipage}[t]{0.235\hsize}
      \centering
      \includegraphics[width=1.0\hsize]{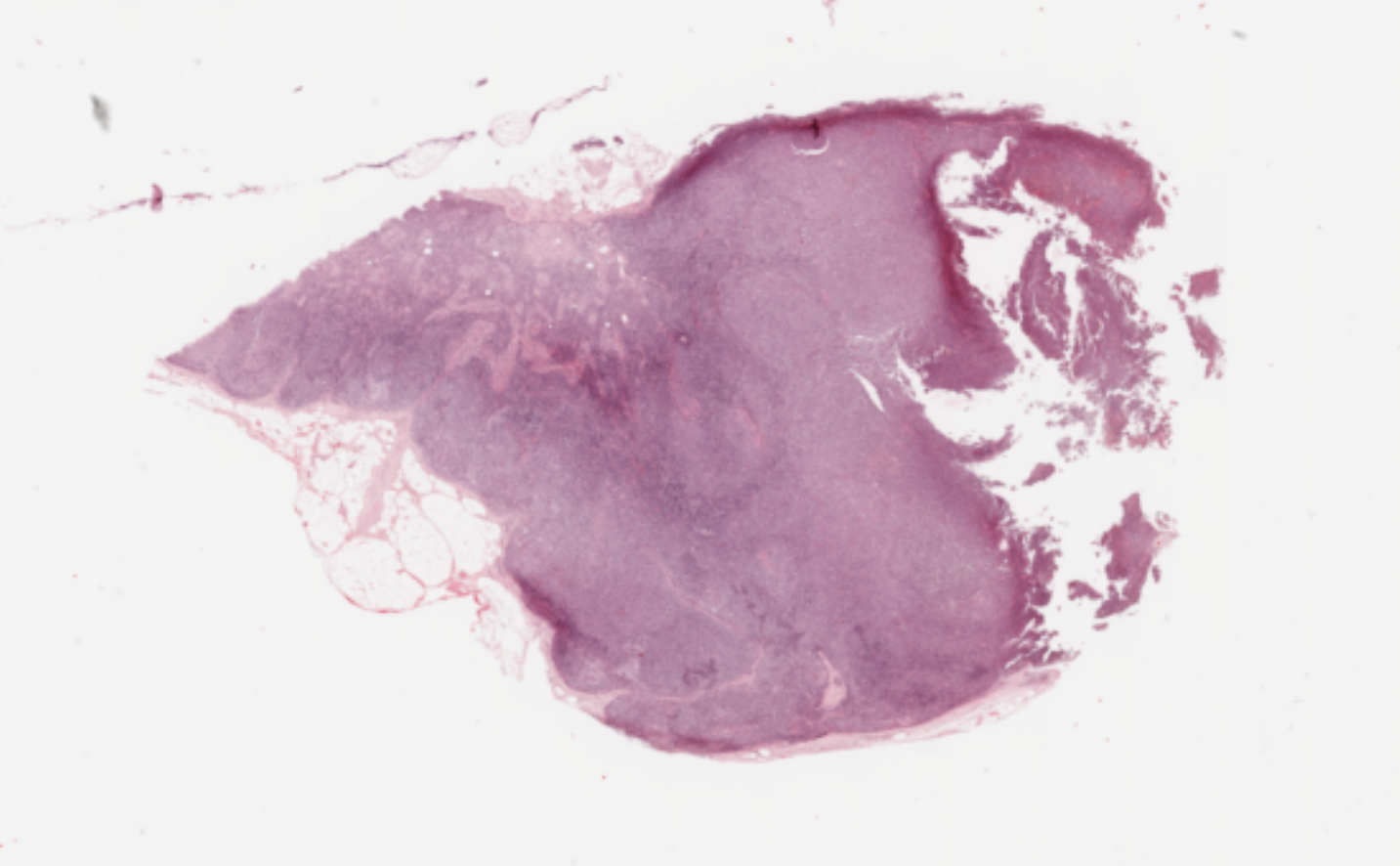}
    \end{minipage} &
    \begin{minipage}[t]{0.235\hsize}
      \centering
      \includegraphics[width=1.0\hsize]{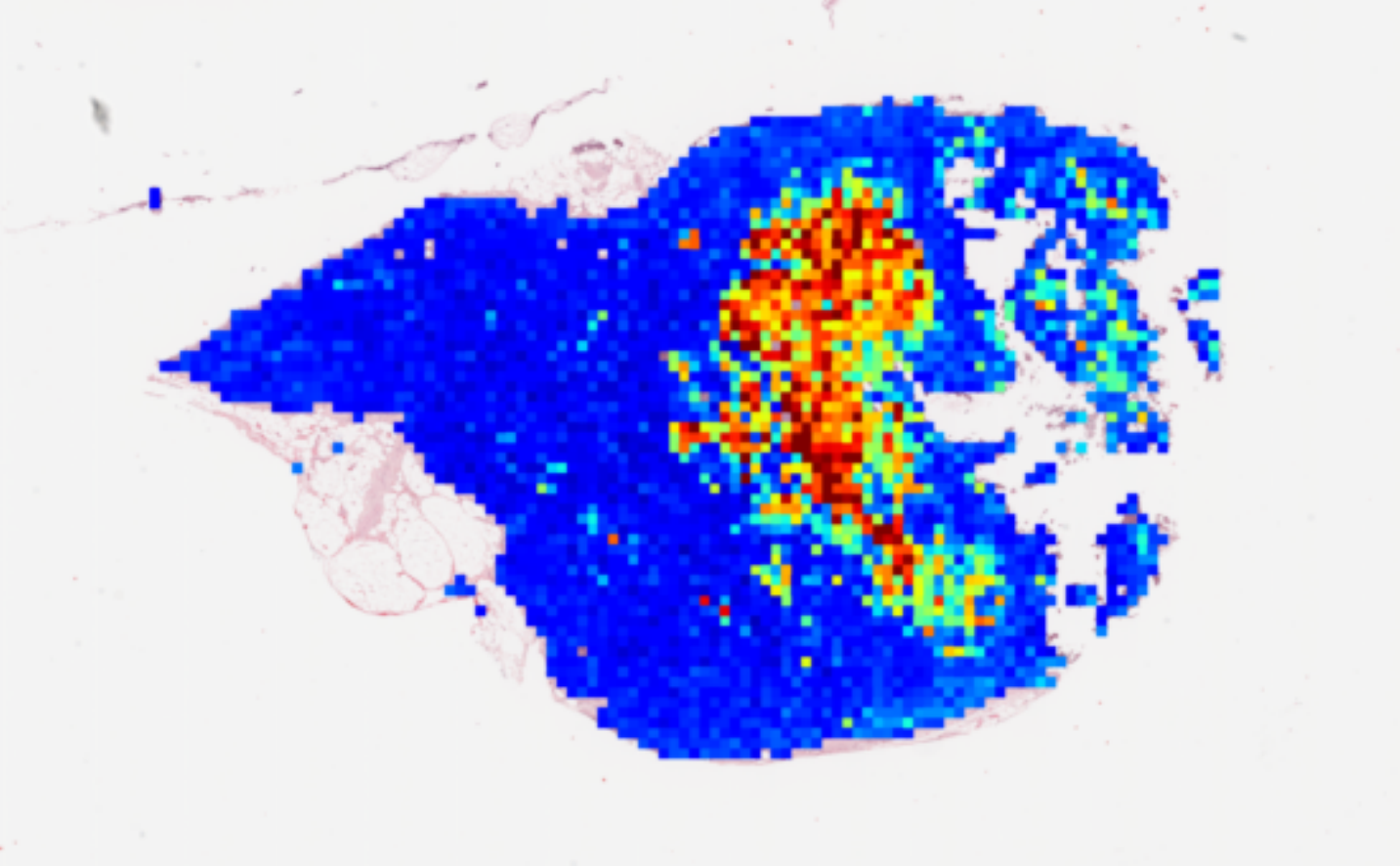}
    \end{minipage} &
    \begin{minipage}[t]{0.235\hsize}
      \centering
      \includegraphics[width=1.0\hsize]{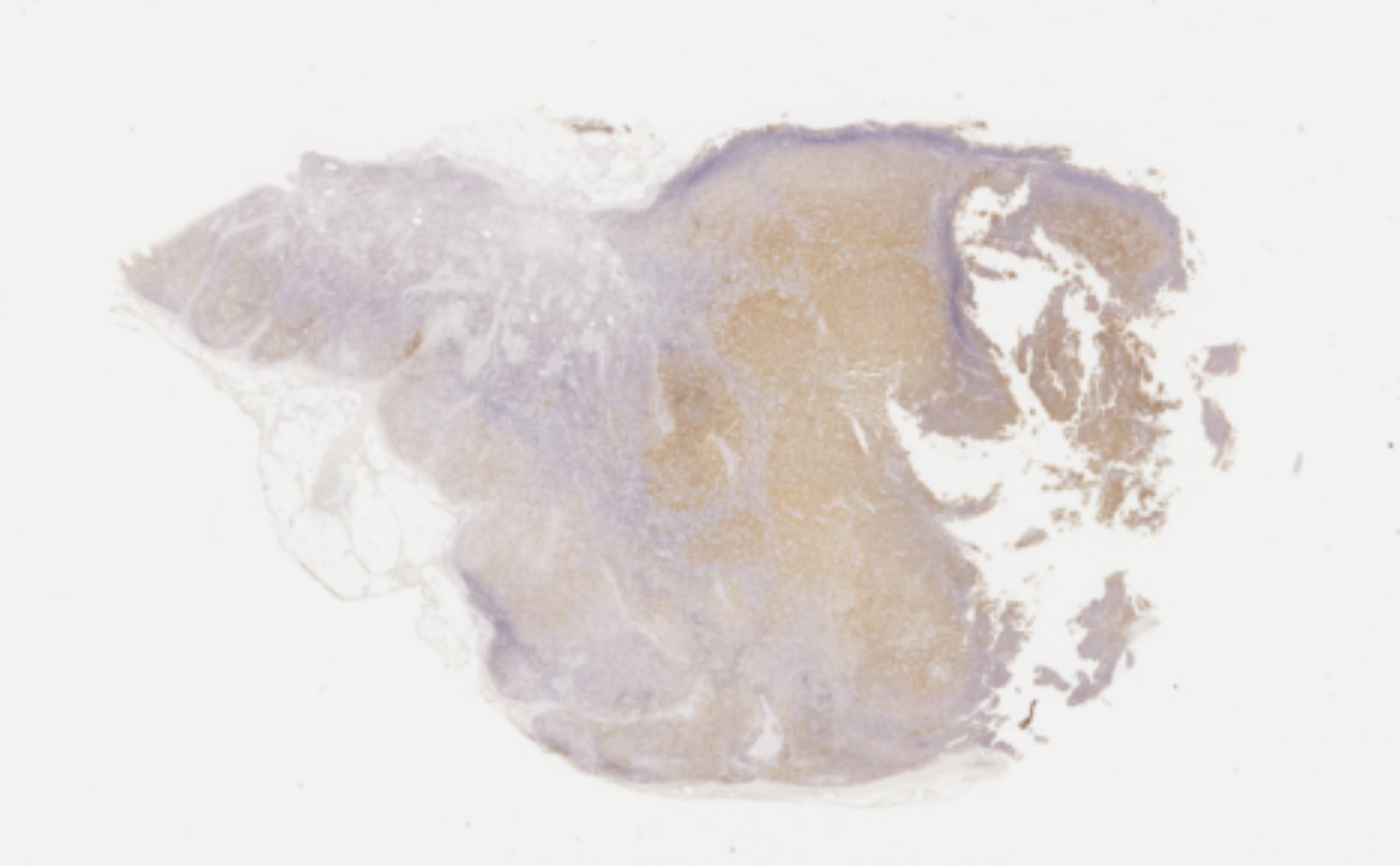}
    \end{minipage}
  \end{tabular}
  \end{center}
  \caption{Visualization of attention weights in DA-MIL and corresponding IHC stained tissues: The left column is original H\&E stained tissue images, the center column is visualized attention weights and the right column is CD20 stained tissue images of the same case. Attention weights in each bag are normalized between 0 to 1, and heat map from blue to red is assigned to between 0 to 1. The attention-weight map in the upper row is generated from 10x WSI, and the lower one is from 20x WSI. We can confirm that the red regions in the visualization results corresponds to the stained regions with brown in CD20 IHC stained tissue specimens.}
  \label{fig:AW}
\end{figure*}

\section{Experiments}

\paragraph{Dataset}
Our experimental database of malignant lymphoma was composed of 196 clinical cases, which represented difficult lymphoma cases from 80 different institutions, and had been sent to an expert pathologist for diagnostic consultation.
As malignant lymphoma has a lot of subtypes, in addition to observing an H\&E stained tissue, serial sections from the same patient's sample are immunohistochemically stained to confirm its expression patterns for final decision making.
It is expected that predicting various subtypes of malignant lymphoma is quite difficult by analyzing only the H\&E stained tissue images and its difficulty was not revealed.
We use the cases with only five typical types of malignant lymphoma: diffuse large B-cell lymphoma (DLBCL), angioimmunoblastic T-cell lymphoma (AITL), classical Hodgkin's lymphoma mixed cellularity (HLMC) and classical Hodgkin's lymphoma nodular sclerosis (HLNS).
In addition, DLBCL is classified into two subtypes; germinal center B-cell (GCB) and non-germinal center B-cell (non-GCB) types.
In this experiment, as a first step, we perform two-class classification, which discriminates DLBCL consisting both GCB and non-GCB types from the other three non-DLBCL classes including AITL, HLMC and HLNS.
In applying our proposed method to this classification problem, DLBCL and non-DLBCL are respectively defined as positive and negative classes, as explained in the previous sections.
Here, the positive instance means that an instance has the capability to discriminate DLBCL from non-DLBCL, and it should be in tumor regions of DLBCL cases because non-tumor regions in DLBCL are expected to have similar features to those in non-tumor regions in non-DLBCL cases.
Hence, a bag indicates a set of image patches extracted from a WSI, where positive instances represent images from tumor regions in DLBCL and negative instances represent images from non-tumor regions in DLBCL and all patches in non-DLBCL.
As the total number of DLBCL cases was 98, the same number of non-DLBCL cases were selected from AITL, HLMC and HLNS cases.
All glass slides of the H\&E stained tissue specimens collected as mentioned above were scanned with an Aperio ScanScopeXT (Leica Biosystems, Germany) at 20x magnification (0.50 um/pixel).

\paragraph{Experimental setup}
In the experiments, we used 10x (1.0 um/pixel) and 20x-magnification (0.50 um/pixel) images, that is, the scale parameter $S$ was set to 2.
We split the dataset mentioned above into 60\% training data, 20\% validation data and 20\% test data, with consideration of patient-wise separation.
In order to generate bags, 100 of $224 \times 224$-pixel image patches were randomly extracted from tissue regions in a WSI for each scale.
The maximum number of bags generated from each WSI was determined as 50.
In extracting image patches for multi-scale, the same regions were selected for each scale as shown in Fig.~\ref{fig:GA}, and we obtained a total of 200 image patches of 100 regions for each bag in our experiment.
In the case where the total number of image patches included in a WSI was less than 3,000, data augmentation was performed by rotating image patches by 90, 180 and 270 degrees.
In the training step, the network trained a bag and renewed parameters for one iteration, and training was performed in 10 epochs where image patches in bags were shuffled for each training epoch.
The domain-regularization parameter $\lambda$ was determined for each epoch as
%
$\lambda = \frac{2}{1 + \exp (-10 r)}-1$,
with
$r = \frac{\text{Current epoch}~m}{\text{Total epochs}~M} \times \alpha$,
%
where $\alpha$ is a hyperparameter, where the best parameter $\alpha$ that showed the highest accuracy on the validation data was set for testing.
In this experiment, VGG16~\cite{simonyan2014very} pre-trained with ImageNet was employed as the feature extractor $G_{\rm{f}}(\cdot; \theta_{\rm{f}})$ and the dimension of the output features was $Q=25,088$.
In the label predictor $G_{\rm{y}}(\cdot; \theta_{\rm{y}})$, a 25,088-dimensional vector was converted into a 512-dimensional vector by the fully connected layer, before the attention mechanism, namely $Q^{\prime}$ was set to 512.
In the attention network, the numbers of input and hidden units were 512 and 128, respectively.
For the domain predictor $G_{\rm{d}}(\cdot; \theta_{\rm{d}})$, a 25,088-dimensional vector was reduced to a 1,024-dimensional vector by the fully connected layer, and a domain label was predicted from it.
The variety of staining conditions could have occurred even if the slides were produced at the same institution, so we regarded each patient as an individual domain in DA learning, and assigned different domain labels to each slide.
Parameters in the network were optimized by SGD momentum~\cite{qian1999momentum}, where the learning rate and momentum were set to 0.0001 and 0.9, respectively.

\begin{figure*}[t]
  \begin{center}
  \begin{tabular}{ccc}
    \begin{minipage}[t]{0.235\hsize}
      \centering
      \includegraphics[width=1.0\hsize]{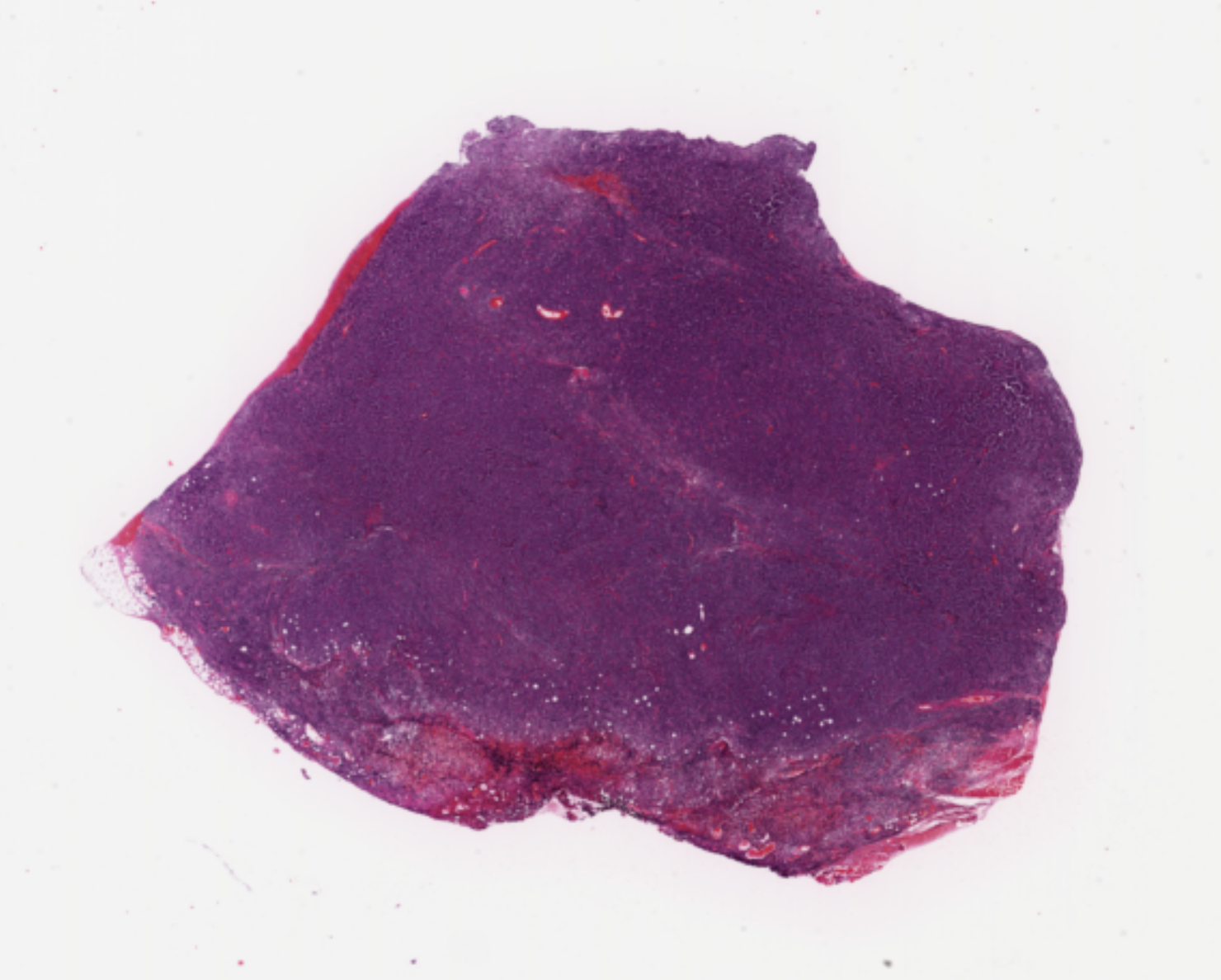}
    \end{minipage} &
    \begin{minipage}[t]{0.235\hsize}
      \centering
      \includegraphics[width=1.0\hsize]{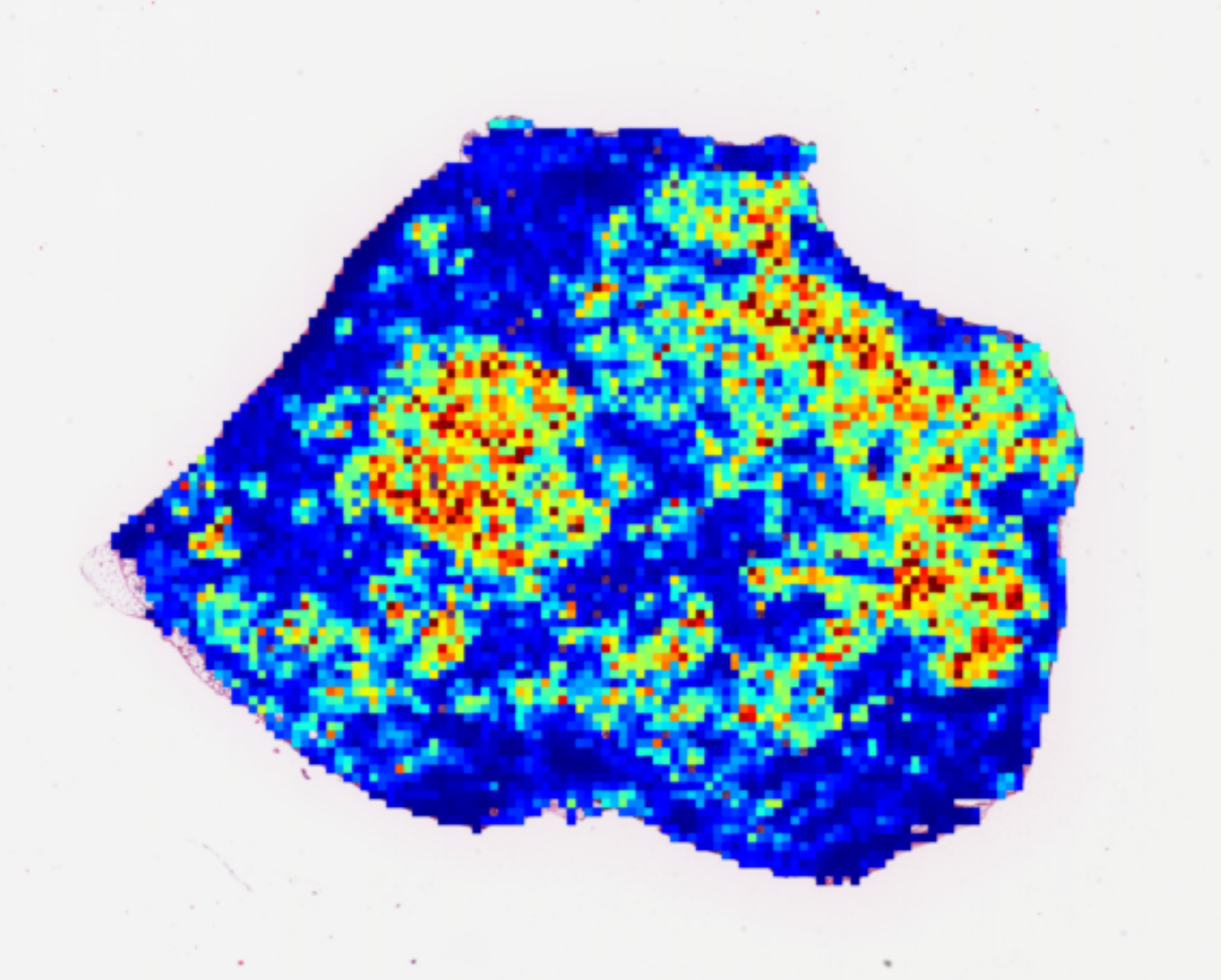}
    \end{minipage} &
    \begin{minipage}[t]{0.235\hsize}
      \centering
      \includegraphics[width=1.0\hsize]{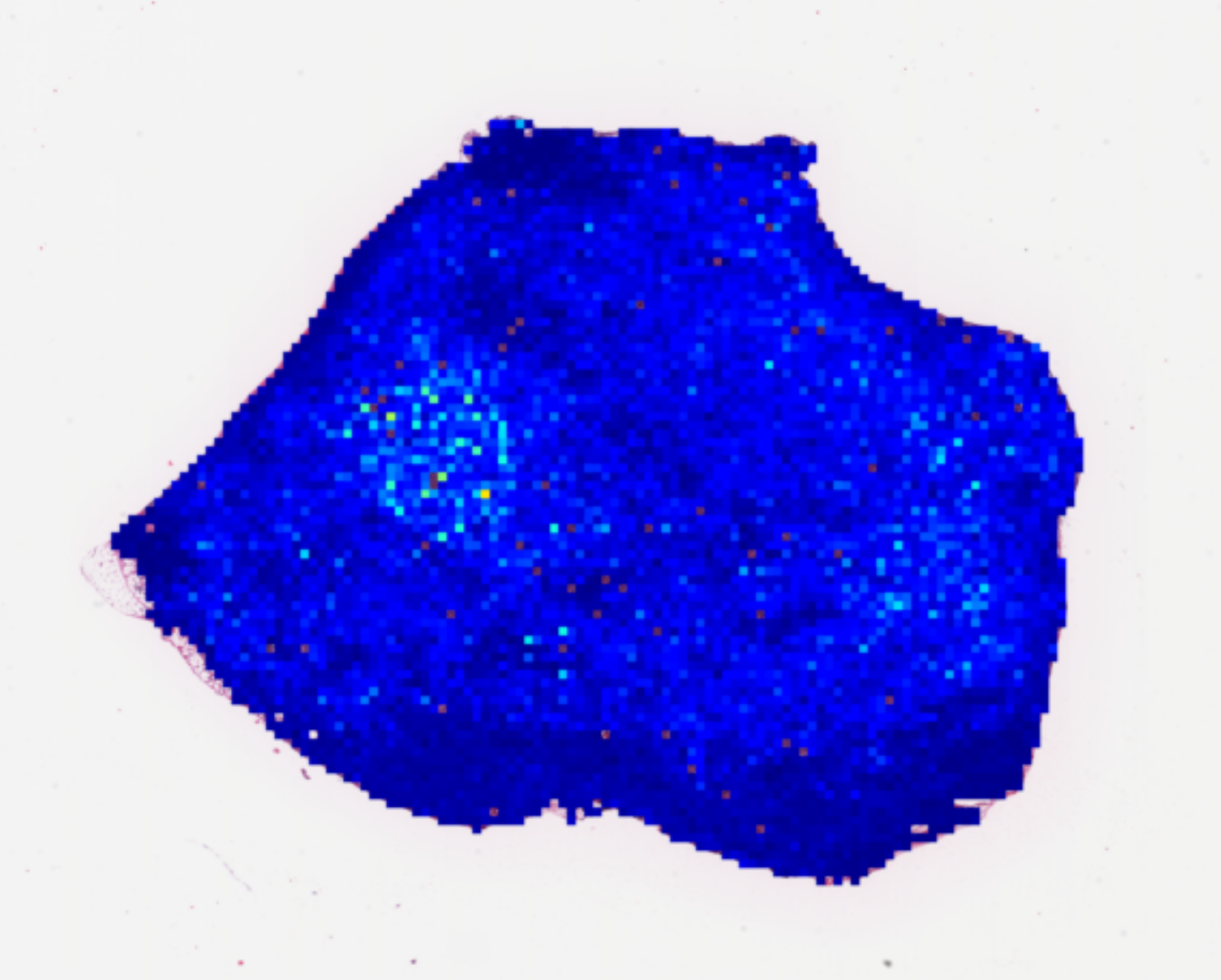}
    \end{minipage}
  \end{tabular}
  \begin{tabular}{ccc}
    \begin{minipage}[t]{0.235\hsize}
      \centering
      \includegraphics[width=1.0\hsize]{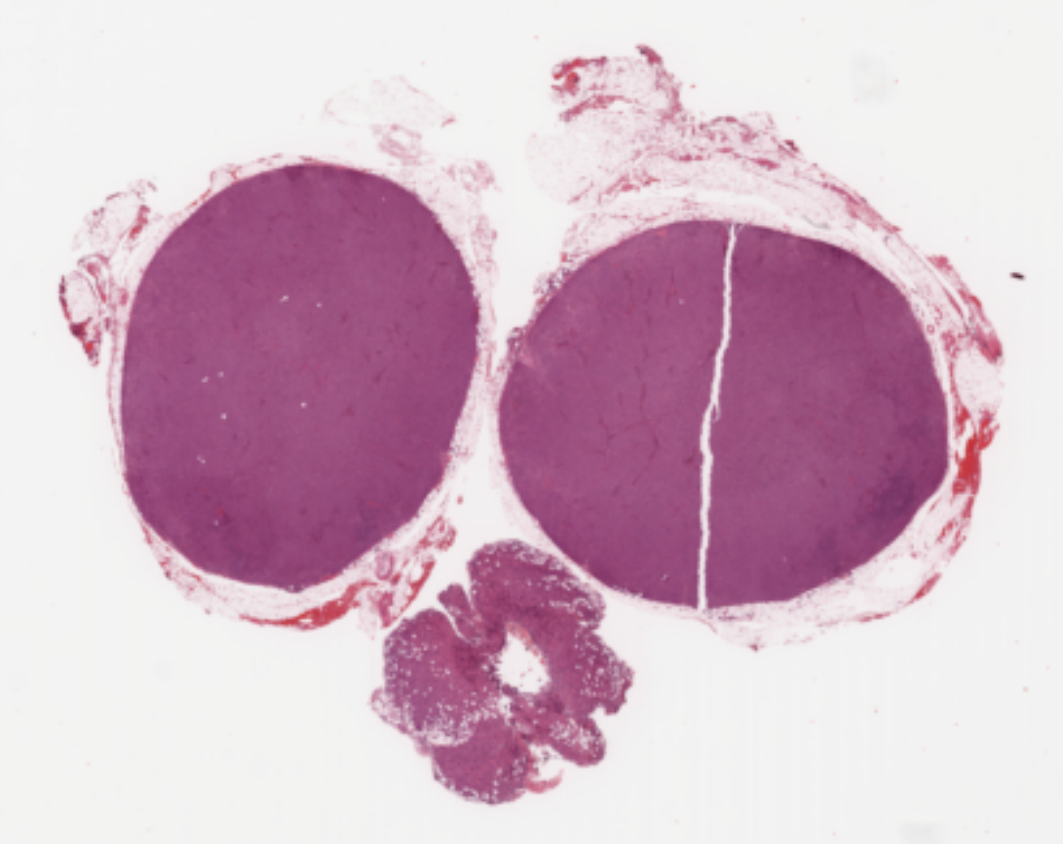}
    \end{minipage} &
    \begin{minipage}[t]{0.235\hsize}
      \centering
      \includegraphics[width=1.0\hsize]{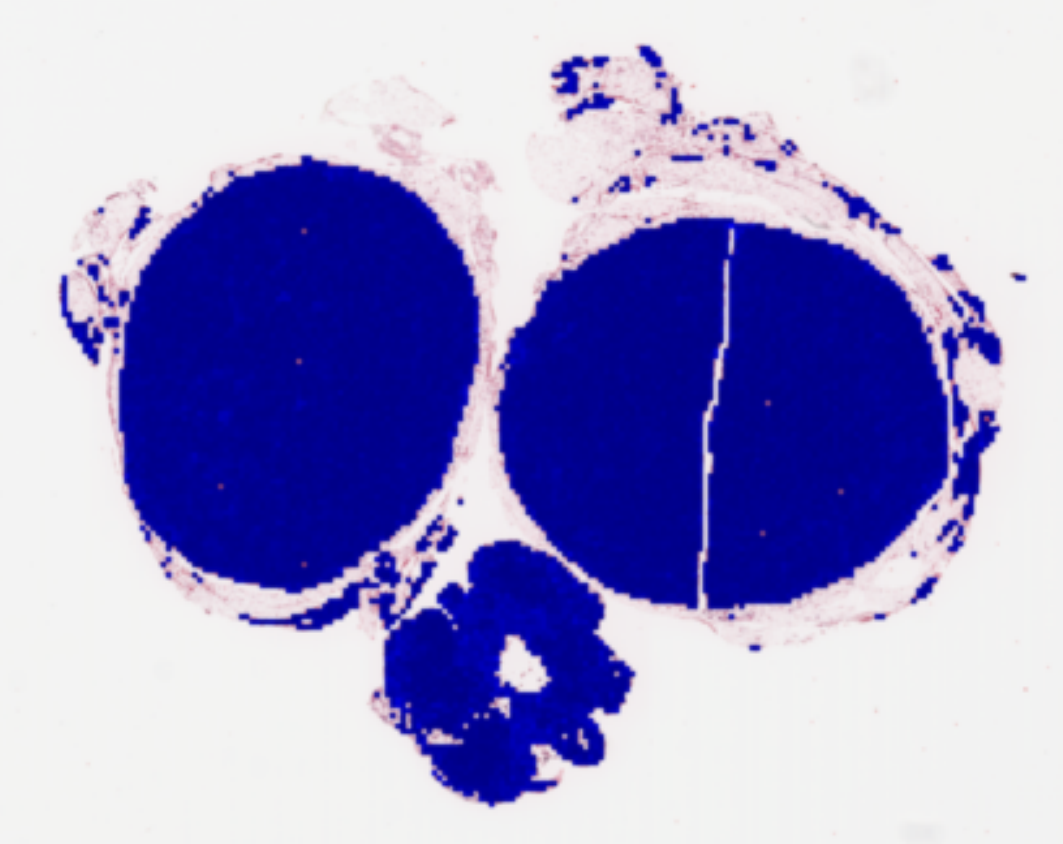}
    \end{minipage} &
    \begin{minipage}[t]{0.235\hsize}
      \centering
      \includegraphics[width=1.0\hsize]{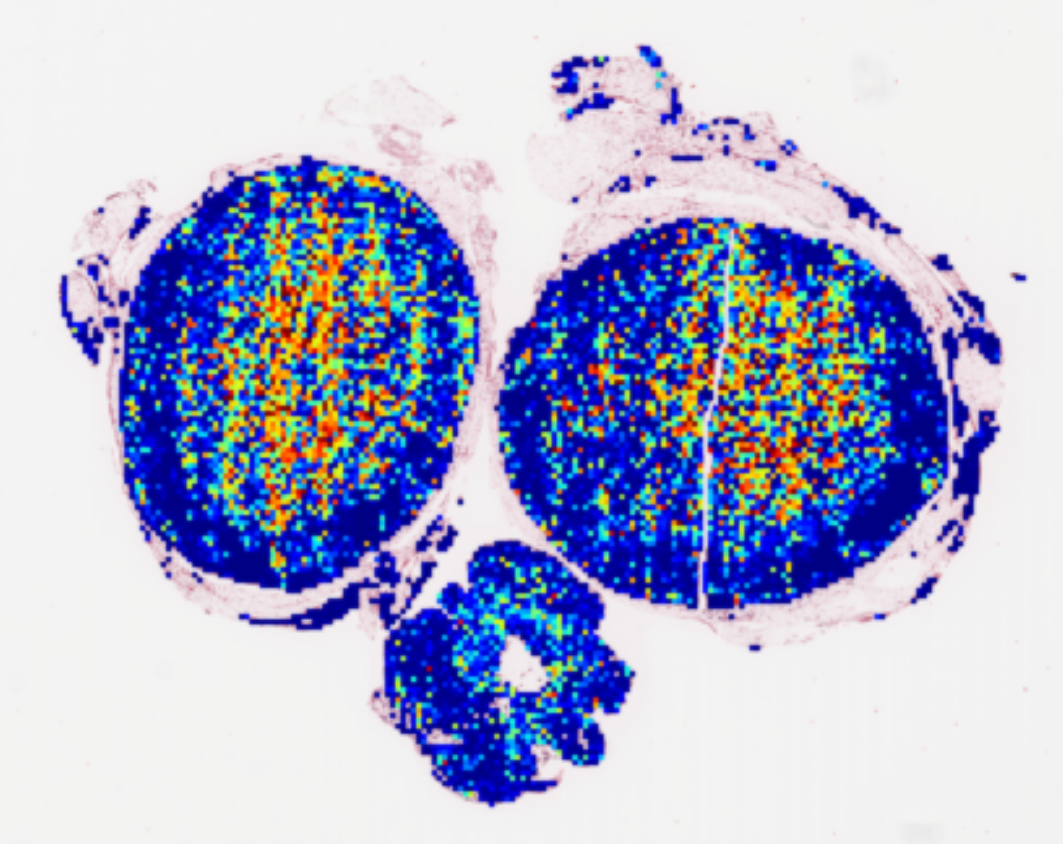}
    \end{minipage}
  \end{tabular}
  \end{center}
  \caption{Visualization of attention weights in MS-DA-MIL inputs: The left column is the original H\&E stained tissue images, and the center and right-hand columns are the visualized attention weights for 10x and 20x by MS-DA-MIL, respectively. We can confirm that one scale had a higher contribution for classification than the other, which means that class-specific features exist at different magnification scales depending on the individual cases.}
  \label{fig:AW_m}
\end{figure*}

  \begin{table*}[t]
      \caption{Comparison of the validation measurement among conventional and proposed methods at each magnification scale, where each result shows the mean value and standard error determined by 5-fold cross validation. Patch-based, attention-based MIL, DA-MIL and MS-DA-MIL were compared, and our proposed method MS-DA-MIL showed the highest accuracy.}
    \begin{center}
      \begin{tabular}{lccccc}
    \hline
    Method & Magnification & Accuracy & Precision & Recall \\ \hline \hline
    \multicolumn{1}{l}{Patch-based} & 10x & 0.740$\pm$0.030 & 0.812$\pm$0.054 & 0.641$\pm$0.049 \\
    \multicolumn{1}{l}{Patch-based} & 20x & 0.754$\pm$0.023 & 0.799$\pm$0.033 & 0.692$\pm$0.057 \\
    \multicolumn{1}{l}{Attention-based MIL} & 10x & 0.811$\pm$0.018 & 0.860$\pm$0.046 & 0.772$\pm$0.071 \\
    \multicolumn{1}{l}{Attention-based MIL} & 20x & 0.826$\pm$0.022 & 0.909$\pm$0.044 & 0.742$\pm$0.061 \\ \hline
      \multicolumn{1}{l}{DA-MIL (ours)} & 10x & 0.836$\pm$0.012 & 0.927$\pm$0.037 & 0.743$\pm$0.046 \\
    \multicolumn{1}{l}{DA-MIL (ours)} & 20x & 0.857$\pm$0.014 & 0.927$\pm$0.039 & 0.793$\pm$0.061 \\
    \multicolumn{1}{l}{MS-DA-MIL (ours)} & 10x, 20x & \bf{0.871$\pm$0.028} & 0.927$\pm$0.025 & 0.813$\pm$0.066 \\ \hline
    \end{tabular}

\end{center}

  \label{tab:acc}
\end{table*}

\paragraph{Results}
Table \ref{tab:acc} shows the classification results of each method, where the values are the means and standard errors determined by 5-fold cross validation.
In the table, ``patch-based'' indicates a CNN classification method whereby the same corresponding label to a case was given for all image patches extracted from the WSI, and where pre-trained VGG16 was used as a CNN model.
The output probability $P_{\rm{patch}}$ of the patch-based method is defined as
$P_{\rm{patch}}(\hat{\mathbb{Y}}_{n}=1) =
 p_{1\_{\rm{patch}}}/(p_{1\_{\rm{patch}}} + p_{0\_{\rm{patch}}})$,
where
\begin{align*}
 p_{1\_{\rm{patch}}} = \exp \left(\frac{1}{\left|\mathcal{I}_n \right|} \sum_{i \in \mathcal{I}_{n}} \log P\left(\hat{y}_{i}=1\right)\right),\\
 p_{0\_{\rm{patch}}} = \exp \left(\frac{1}{\left|\mathcal{I}_{n}\right|} \sum_{i \in \mathcal{I}_{n}} \log P\left(\hat{y}_{i}=0\right)\right).
\end{align*}
Here, $\mathcal{I}_n$ is a set of image patches extracted from the $n^{\rm{th}}$ WSI, and $P(\hat{y_{i}}=1)$ is the probability for an input image patch $\bm{x}_i$ to be classified to DLBCL.
The maximum number of $224 \times 224$-pixel image patches extracted from each WSI was set to 5,000, as the case had instances from the same number of regions.
DA-MIL in the table has the same meaning as MS-DA-MIL with scale parameter $S=1$.
We confirmed that MS-DA-MIL showed the highest classification accuracy compared with those of patch-based and attention-MIL.
In particular, it was confirmed that the classification accuracy of MS-DA-MIL was higher than that of DA-MIL, which could provide us with encouragement to make use of multi-scale input for pathology image classification.

In addition, we visualized the distribution of attention weights, which were calculated for correctly classified cases into DLBCL.
Figure~\ref{fig:AW} shows the images of an H\&E stained tissue, corresponding attention-weight map and CD20 immunohistochemically stained tissue specimen for a serial section of the same case.
For the attention-weight maps in the middle columns of Fig.~\ref{fig:AW}, attention weights were normalized between 0 to 1 in each bag, and blue to red (0 to 1) heat map was generated.
Thus, red regions in the attention-weight maps represent the highest contribution for classification in each bag.
Because CD20 is an IHC staining that neoplastic B-cells mainly react and shows strong positivity, we can visually confirm the validity of the attention weights of the proposed DA-MIL.
In CD20 stained images, positive regions are stained in brown by diaminobenzidine and negative regions are stained in blue by hematoxylin.
In comparison to those images, we can see that the attention weights showed higher values in the CD20-positive regions.
On the other hand, CD20-negative regions showed low values in the attention-weight maps, and image patches in such regions did not contribute to classification.
According to the above results, we showed the appropriate assignment of attention weights.
Figure~\ref{fig:AW_m} shows the images of an H\&E image and its attention-weight maps calculated by MS-DA-MIL.
Similarly to DA-MIL, attention weights in each bag were normalized from 0 to 1, and attention-weight maps for each scale were generated with heat map.
As we can see in Fig.~\ref{fig:AW_m}, one of them has high attention weights in the 10x image, while the other shows high attention weights in the 20x image.
Therefore, there exists appropriate magnification to obtain class-specific features depending on the cases, and MS-DA-MIL could consider this and show the effectiveness of multi-scale input analysis.

\section{Conclusion}
We proposed a new CNN for cancer subtype classification from unannotated histopathological images which effectively combines MI, DA, and MS learning frameworks in order to mimic the actual diagnosis process of pathologists.
When the proposed method was applied to malignant lymphoma subtype classifications of 196 cases, the performance was significantly better than that of standard CNN or other conventional methods, and the accuracy compared favorably with that of standard pathologists.

 \section*{Acknowledgment}
 This work was partially supported by Hori Sciences and Arts Foundation to N.H., S.N., H.H. and I.T., MEXT KAKENHI 26108003 to H.H., 17H00758, 16H06538 to I.T., JST CREST JPMJCR1502 to I.T. and RIKEN Center for Advanced Intelligence Project to I.T.

\clearpage

{\small
\bibliographystyle{ieee}
\bibliography{ref}
}

\end{document}